\pdfoutput=1

\documentclass[11pt]{article}

\usepackage[final]{acl}

\usepackage{times}
\usepackage{latexsym}

\usepackage[T1]{fontenc}

\usepackage[utf8]{inputenc}

\usepackage{microtype}

\usepackage{inconsolata}

\usepackage{graphicx}

\usepackage{amsmath}
\usepackage{amsfonts}
\usepackage{graphicx}
\usepackage{subfigure}
\usepackage{booktabs}       
\usepackage[ruled,linesnumbered]{algorithm2e}               
\usepackage{multirow}
\usepackage{tcolorbox}
\usepackage{xcolor}

\title{Unlocking the Planning Capabilities of Large Language Models with Maximum Diversity Fine-tuning}

\author{Wenjun Li, Changyu Chen, Pradeep Varakantham \\
  Singapore Management University \\
  \texttt{\{wjli.2020, cychen.2020, pradeepv\}@smu.edu.sg}\\
}

\begin{document}
\maketitle

\begin{abstract}
Large language models (LLMs) have demonstrated impressive task-solving capabilities through prompting techniques and system designs, including solving planning tasks (e.g., math proofs, basic travel planning) when sufficient data is available online and used during pre-training. However, for planning tasks with limited prior data (e.g., blocks world, advanced travel planning), the performance of LLMs, including proprietary models like GPT and Gemini, is poor. This paper investigates the impact of fine-tuning on the planning capabilities of LLMs, revealing that LLMs can achieve strong performance in planning through substantial (tens of thousands of specific examples) fine-tuning. Yet, this process incurs high economic, time, and computational costs for each planning problem variation. To address this, we propose Clustering-Based Maximum Diversity Sampling (CMDS), which selects diverse and representative data to enhance sample efficiency and the model's generalization capability. Extensive evaluations demonstrate that CMDS-$l$, a baseline method combining CMDS with language embeddings, outperforms random sampling. Furthermore, we introduce a novel algorithm, CMDS-$g$, which encodes planning task instances with their graph representations into the embedding space. Empirical results show that CMDS-$g$ consistently outperforms baseline methods across various scales and multiple benchmark domains.
\end{abstract}

\section{Introduction}
System 1 competencies are characterized by fast, instinctive, and emotional responses, while System 2 competencies involve slower, more deliberate, and logical thinking processes \cite{Kahneman:2011fj}. Prior studies~\cite{valmeekam2024planning,pallagani2023understanding,liu2023llm+,guan2023leveraging,kambhampati2024llms} have argued that LLMs struggle to generate valid plans in the automated planning domain due to weak System 2 competencies, despite demonstrating impressive planning capabilities in tasks such as Minecraft \cite{wang2024describe,yuan2023skill} and household planning \cite{huang2022language,yao2022react}, where their System 1 competencies are more relevant (due to availability of large datasets online). Our study demonstrates that LLMs can indeed achieve System 2 competencies through fine-tuning with sufficiently large datasets.

This paper focuses on enhancing the planning capabilities of LLMs in rigorous, automated settings—termed System 2 planning. We distinguish this from System 1 planning, which covers tasks in environments like Minecraft and household activities. There are two key differences between these types of planning: 1) Complexity: System 2 planning is significantly more complex than System 1 planning, requiring deeper reasoning and more steps to solve; 2) Data availability: The amount of pre-training data available for System 1 tasks far exceeds that for System 2 tasks. For instance, the online data available for completing household tasks is much greater than that for finding optimal solutions to complex logistics problems.

Due to these challenges, LLMs initially exhibit weaknesses in automated planning (System 2) and researchers have been focusing on designing prompts and pipelines. However, these techniques are proven to have little improvement in improving LLM's capabilities in automated planning \cite{valmeekam2024planning,valmeekam2024planbench}. A detailed related work section is provided in Appendix \ref{app:related_work}. In this paper, we employ fine-tuning as a targeted approach and demonstrate that fine-tuning with sufficiently large datasets enables LLMs to achieve robust planning capabilities. Moreover, we provide an in-depth analysis of how different factors, such as data scaling, diversity, and complexity, impact the fine-tuning outcomes. We specifically examine how these factors influence both the planning capabilities and generalization performance of LLMs. By addressing the gap in the current research, our study offers valuable insights into how to fine-tune LLMs more effectively and efficiently to enhance their planning competencies.

\begin{figure}[t]
  \centering
  \includegraphics[width=1.0\linewidth]{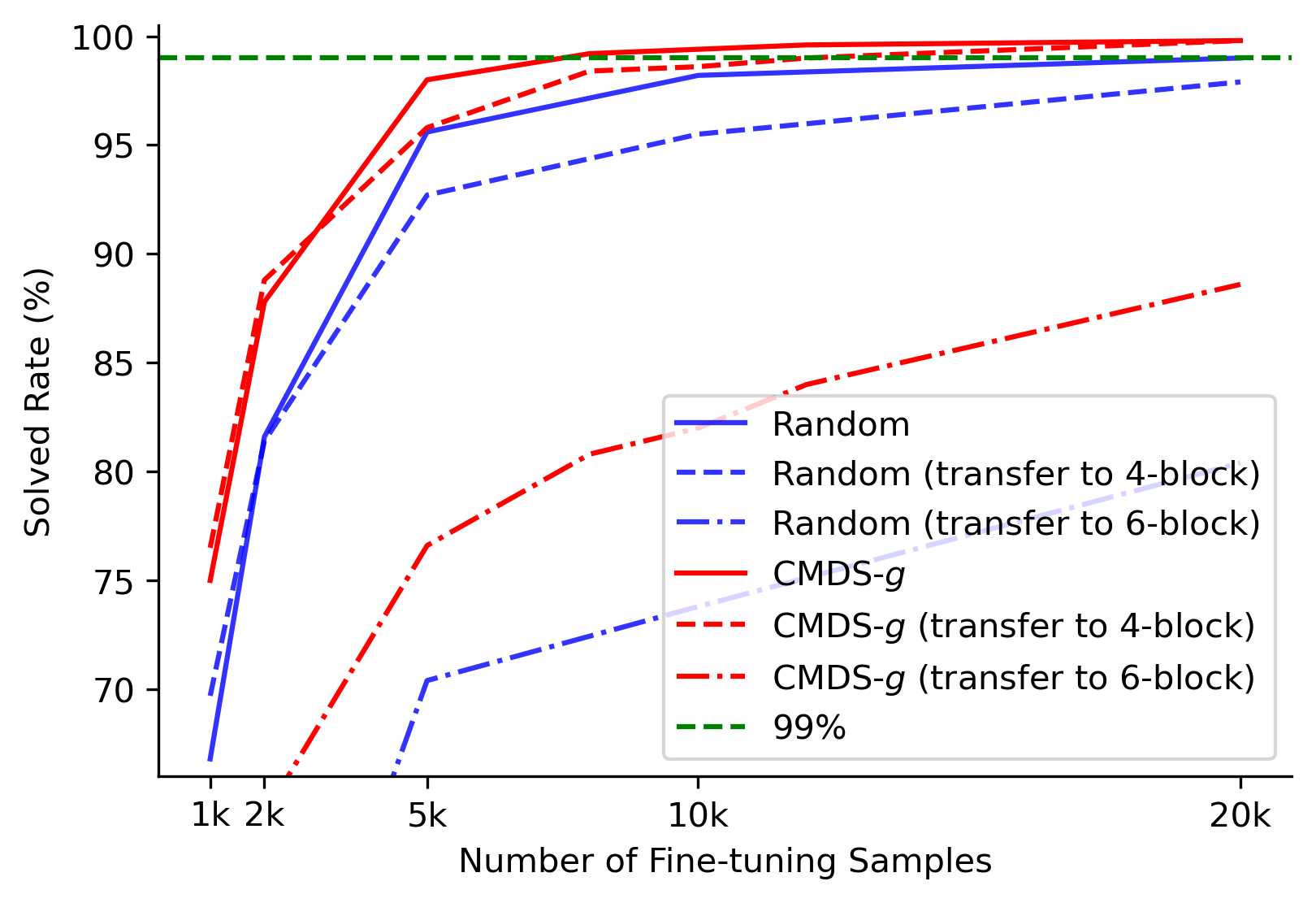}
  \caption{Scaling effect of fine-tuning \texttt{Llama-3-8b} in Blocksworld. Solved rate on hold-out (5-block) and transfer (4-block and 6-block) testing. A clearer comparison is in Appendix \ref{app:data_scaling}.}
  \label{fig:massive_fine_tuning}
\end{figure}

\begin{figure*}[ht]
  \centering
  \includegraphics[width=1.0\linewidth]{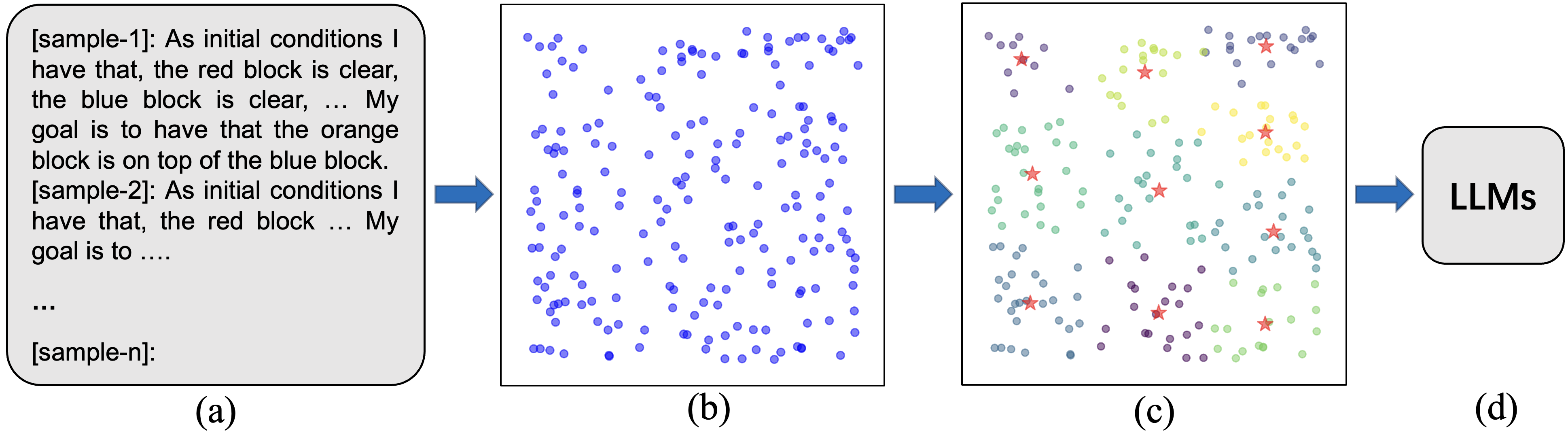}
  \caption{Overview of CMDS. (a) Training samples in natural language. (b) Samples encoded in embedding space. (c) A subset of samples selected by CMDS. The red stars are the selected ones. (d) Fine-tune LLMs with the subset of samples.}
  \label{fig:algo_pipeline}
\end{figure*}

Fine-tuning both closed-source models like OpenAI's GPT-x~\cite{brown2020language} and open-source models such as Meta's Llama-x~\cite{touvron2023llama} entails substantial time, economic, and computational costs. For instance, fine-tuning \texttt{GPT-3.5-turbo} on just 10,000 instances in Blocksworld (domain details provided in Experiment Setup) costs approximately 1,200 USD. Similarly, fine-tuning \texttt{Llama-3-8b} on this amount of data requires high-end GPUs and can take hundreds of GPU hours. Given these constraints, sample efficiency in fine-tuning is particularly crucial, especially for applications where models need frequent updates or rapid adaptation to new domains and complex tasks that require huge amounts of fine-tuning data. For example, housekeeping robots must efficiently adapt to various home layouts and furniture arrangements.

This paper aims to enhance the planning capabilities of LLMs while maximizing the sample efficiency and the model's generalization capability. We introduce a simple yet effective approach called Clustering-Based Maximum Diversity Sampling (CMDS), which selects diverse and representative data in the embedding space. Moreover, we propose CMDS-$g$ to encode planning instances with their graph representations into the embedding space. A motivating example is illustrated in Figure \ref{fig:massive_fine_tuning}. We generated tens of thousands of 5-block instances in Blocksworld and applied both Random sampling and CMDS-$g$ to select fine-tuning samples. CMDS-$g$ consistently outperforms Random at the same scale and demonstrated much higher sample efficiency as the data size increased. To achieve a 99\% solved rate, Random sampling requires around 20,000 samples, whereas CMDS-$g$ needs only about 7,500, effectively reducing the training time and computational costs by more than half (from 40 to 15 A100 GPU hours in our experiment). More importantly, CMDS-$g$ exhibits superior generalization capabilities when the fine-tuned model is transferred to 4-block and 6-block test instances. These advantages become even more critical for large-scale planning tasks, where the fine-tuning data required to achieve satisfactory performance grows exponentially with task complexity. For example, exponentially more samples are needed for LLMs to perform equivalently on 6-block tasks as on 5-block tasks.

In summary, our research offers three key contributions. First, we provide a comprehensive analysis of how data scaling, diversity, and task complexity influence the fine-tuning outcomes of LLMs, particularly in automated planning. Our findings show that LLMs can achieve System 2 competencies through fine-tuning with sufficiently large datasets. Second, we introduce CMDS, a simple yet effective method for selecting diverse and representative data in the embedding space, thereby enhancing the sample efficiency of fine-tuning. Third, we propose CMDS-$g$, a novel approach that leverages graph representations to encode planning instances. Extensive experiments demonstrate that CMDS-$g$ significantly boosts sample efficiency and generalization, consistently outperforming existing baseline methods.

\begin{figure}[ht]
  \centering
  \includegraphics[width=1.0\linewidth]{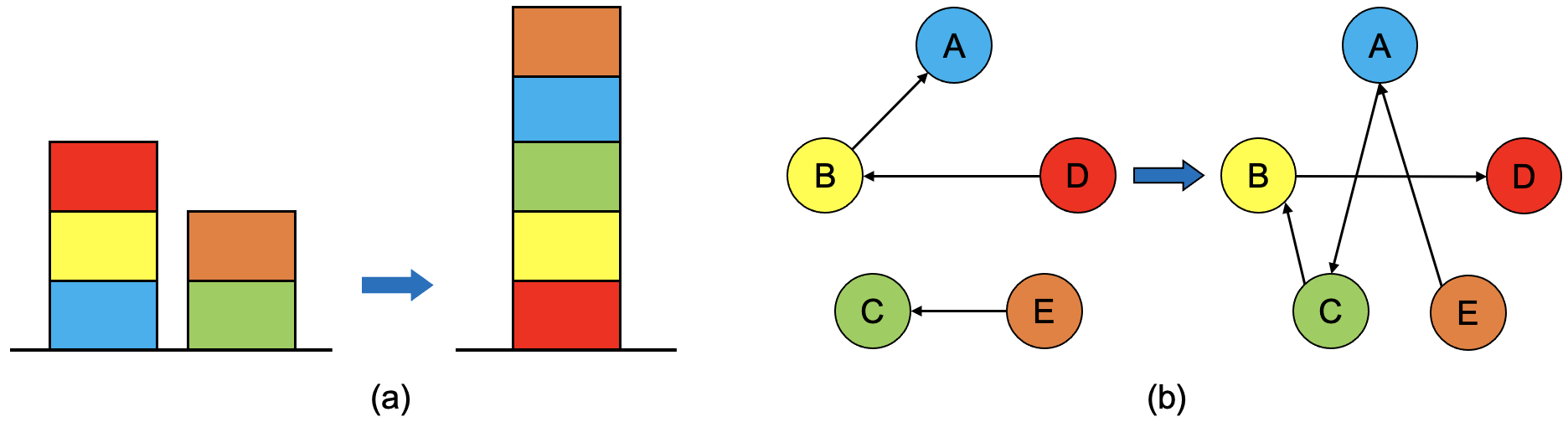}
  \caption{(a) An example task from Blocksworld. Left: initial configuration. Right: goal configuration. (b) A graph representation of the example task, where nodes are objects and edges are predicates.}
  \label{fig:blocksworld_domain}
\end{figure}

\section{Clustering-Based Maximum Diversity Sampling} \label{sec:approach}
Fine-tuning data consists of two components: queries and the associated responses. Previous studies~\cite{zhou2024lima,zhao2024long} have shown that higher quality responses and more diverse queries lead to better LLM performance after fine-tuning. In automated planning, we can achieve the highest quality in responses by employing traditional planning solvers (e.g., Fast Downward~\cite{helmert2006fast}) to find the optimal plans. As a result, how to capture the diversity of queries in fine-tuning data becomes the main challenge. 

Unlike previous work, which relied on human resources to capture diversity in queries, we aim to automatically identify diverse and effective samples to achieve higher sample efficiency. The task of maximizing diversity within a subset of samples is known as the Maximum Diversity Problem (MDP,~\cite{ghosh1996computational,marti2013heuristics}), a well-studied NP-hard problem. This means that finding the exact optimal solution is computationally challenging for large datasets. In this paper, we introduce a simple yet effective method called Clustering-Based Maximum Diversity Sampling (CMDS) to identify representative samples, thereby ensuring diversity within the dataset.

\subsection{Representing Planning Tasks as Graphs}
MDP algorithms operate in numeric vector spaces, so the first step is to convert planning tasks from natural language into the embedding space.

Natural language can be efficiently encoded into the embedding space using tools such as Universal Sentence Encoder~\cite{cer2018universal} and Sentence-BERT~\cite{reimers2019sentence}. However, due to the narrative nature of planning tasks, tasks with distinct solutions can only differ in a few words in the natural language descriptions and result in highly similar language embeddings. For example, in Figure \ref{fig:blocksworld_domain}(a), switching the positions of the orange and green blocks in the initial configuration while keeping other conditions fixed changes the task descriptions slightly. The original task description is, ``As initial conditions, I have that the orange block is clear, the orange block is on top of the green block, ...", whereas the altered task description is, ``As initial conditions, I have that the green block is clear, the green block is on top of the orange block, ...". Despite having different solutions, these two tasks produce very similar language embeddings that are close in vector space. This issue makes subsequent diversity maximization ineffective. Additional results and analyses on the limitations of language embeddings are presented in Appendix \ref{app:language_embeddings}.

Therefore, a more representative embedding approach specifically tailored for planning tasks is necessary to capture the nuances between tasks. Inspired by ~\cite{rivlin2020generalized,silver2021planning}, we use graph representations to encode planning tasks. Each planning task consists of an initial configuration and a goal configuration of objects. To solve the task, LLMs need to perform sequential actions to transform the initial configuration into the goal configuration. In the graph representation, each object is a node, and each predicate is a (directed) edge between two nodes. An example task from Blocksworld and its graph representation is shown in Figure \ref{fig:blocksworld_domain}. Once the planning tasks are converted into graphs, we can fully capture the task information and efficiently encode these graphs into vector embeddings by concatenating the vectors of the initial configuration and the goal configuration into a single graph representation. Details are provided in Appendix \ref{app:domain_details}.

\begin{algorithm}[ht]
\caption{CMDS}
\label{alg:select_mdp}
\KwIn{$N$ tasks, subset size $k$, embedding method $f(\cdot)$, distance metric $c(\cdot, \cdot)$}
Initialize empty subset $\mathcal{D}$, embeddings set $\mathbb{E}$ \\
\For{$i=1$ to $N$}{
    Encode task $t_i$ as vector embedding, $e_i = f(t_i)$ \\
    Insert $e_i$ into $\mathbb{E}$ 
}
Dimension reduction in embedding space \\
Perform clustering with distance metric $c(\cdot, \cdot)$ over the embeddings with $k$ clusters \\
Obtain $k$ centroids of the clusters \\
\For{$i=1$ to $k$}{
    In each cluster, find the $e_i$ closest to the cluster centroid \\
    Retrieve the task instance associated with $e_i$ and add it to $\mathcal{D}$ \\
}
\KwOut{A subset of tasks, $\mathcal{D}$}
\end{algorithm}

\subsection{Selecting Maximum Diversity Data}
To enhance the effectiveness of downstream fine-tuning, we aim to select a subset of samples that are not only diverse but also representative of the overall dataset. We introduce a simple yet effective method called Clustering-Based Maximum Diversity Sampling (CMDS), designed to achieve this balance by selecting diverse and representative samples in the embedding space.

Assume we have a total number of $N$ tasks and we want to find a subset of $k$ tasks that maximize the overall diversity, defined as the sum of all pairwise distances between the $k$ samples. To make the clustering more effective, we first perform dimension reduction (e.g., t-SNE~\cite{van2008visualizing}) in the embedding space. Subsequently, we perform k-means clustering on the entire set of data points, with $k$ used as the number of clusters. For each cluster, the data point closest to the cluster centroid is selected to represent the cluster in the final subset. When CMDS is applied with graph embeddings, we refer to it as CMDS-$g$. 

Additionally, we propose an improved baseline method, CMDS-$l$, which combines CMDS with language embeddings. While language embeddings have been widely used to estimate dataset diversity, they have not previously been integrated with an automatic sample selection algorithm. Our extensive experiments demonstrate that CMDS-$l$ outperforms random sampling. Due to its general applicability, CMDS-$l$ is valuable for fine-tuning LLMs in domains beyond automated planning.

Our clustering-based method is compatible with any embedding approaches that can convert planning tasks from natural language to vector representations. The effectiveness of clustering depends on an appropriate distance metric; for example, we used the L2 norm distance for language embeddings and edit distance for graph embeddings in our experiments. The complete procedures of CMDS-$l$ and CMDS-$g$ are detailed in Algorithm \ref{alg:select_mdp}.

\section{Experiment Setup} \label{experiment_setup}
\textbf{Benchmark Domains:} We utilize two widely adopted benchmark domains for automated planning from the International Planning Competition (IPC,~\cite{long20033rd}). Detailed information about domain properties, actions, predicates, dataset generation, prompts, and other specifics can be found in Appendix \ref{app:domain_details}.

Blocksworld: This domain consists of blocks, a table, and a robot hand. Blocks can be placed on other blocks or on the table. A block with nothing on top is clear, and the robot hand can hold one block or be empty. The robot can perform four actions: pick-up, put-down, stack, and unstack. The goal is to transition from the initial to the goal configuration of blocks, uniquely identified by their colors. Task complexity mainly varies with the number of blocks. We created random instances with different block counts, separating training and testing tasks to ensure LLMs do not see testing tasks during fine-tuning. Detailed experimental settings are provided with each table and figure.

Logistics: The goal is to transport packages to specified locations. Locations are grouped by cities, with trucks moving packages within a city and airplanes moving packages between cities. Each city has one truck and an airport, and there can be multiple airplanes and packages. The Logistics domain is much more complex than Blocksworld, requiring longer plans. We generated distinct instances with varying numbers of cities, locations, packages, and airplanes.

\textbf{Base Models:} We use \texttt{GPT-3.5-turbo-0125} and \texttt{Llama-3-8b} for Research Question 1 (RQ-1), and include \texttt{Llama-2-7b} \cite{touvron2023llama} for Research Question 2 (RQ-2). \texttt{GPT-4} \cite{achiam2023gpt} was omitted from our experiments due to its limited experimental access and lack of public availability for fine-tuning. 

\textbf{Baselines:} We compare CMDS-$g$ with two baselines: Random and CMDS-$l$. Random uniformly samples $k$ instances from the entire training dataset. CMDS-$l$ differs from CMDS-$g$ only in the embedding method and distance metric, as described in section \ref{sec:approach} and Algorithm \ref{alg:select_mdp}.

\textbf{Evaluation:} In our experiments, LLMs respond in natural language, and the generated plans are translated to PDDL and then verified using VAL~\cite{howey2004val}. A plan is considered correct as long as it successfully transitions from the initial configuration to the goal configuration. An analysis of the optimality rate of the generated plans is provided in Appendix \ref{app:plan_optimality}.

\textbf{Prompt:} We use two types of prompts: one-shot and zero-shot. The one-shot prompt includes domain instructions, an example task with its solution, and a query task. The zero-shot prompt contains only domain instructions and the query task. Further details about the prompt settings are available in Appendix \ref{app:domain_details}. We found that fine-tuned models perform better with zero-shot prompts because they have seen many examples during fine-tuning, enabling accurate and concise responses. Conversely, un-fine-tuned models perform better with one-shot prompts. Therefore, we evaluate un-fine-tuned models with one-shot prompts and fine-tuned models with zero-shot prompts.

\begin{table*}[ht]
  \centering
  \begin{tabular}{cccccccc}
    \toprule
    & & \multicolumn{3}{c}{\textbf{GPT-3.5-turbo}} & \multicolumn{3}{c}{\textbf{Llama-3-8b}} \\
    \cmidrule(r){3-5} \cmidrule(r){6-8}
    \multicolumn{2}{c}{\textbf{Domain}} & \textbf{Pre-FT} & \multicolumn{2}{c}{\textbf{Post-FT}} & \textbf{Pre-FT} & \multicolumn{2}{c}{\textbf{Post-FT}} \\
    \cmidrule(r){4-5} \cmidrule(r){7-8}
    & & & $k=100$ & $k=1000$ & & $k=100$ & $k=1000$ \\
    \midrule
    \multirow{4}{*}[0em]{\textbf{Blocksworld}} 
      & 3-block & 8.0\% (8/100) & 65.0\% & 98.0\%   & 0.0\% & 57.0\% & 85.0\% \\
      & 4-block & 3.6\% (18/500) & 61.8\% & 91.9\% & 0.0\% & 43.9\% & 91.3\% \\
      & 5-block & 1.2\% (6/500) & 36.4\% & 65.8\%   & 0.0\% & 19.6\% & 46.4\% \\
      & 6-block & 0.4\% (2/500) & 15.4\% & 34.0\%   & 0.0\% & 3.6\%  & 15.0\% \\
    \midrule
    \textbf{Logistics} & & 0.7\% (2/300) &7.5\%  &73.0\%     & 0.0\% &2.8\%  &72.7\% \\
    
    \bottomrule
  \end{tabular}
  \caption{Task solved rate of pre-fine-tuning (Pre-FT) and post-fine-tuning (Post-FT) models in Blocksworld and Logistics domain. $k$ is the number of random samples used in fine-tuning. We use 4-block samples for fine-tuning in Blocksworld and samples with varying numbers of variables for fine-tuning in Logistics. GPT and Llama results are collected across three and five random seeds respectively.}
  \label{table:before_after_finetuning}
\end{table*}

\section{Experiment Results}
\subsection{RQ-1. How does fine-tuning impact the planning capabilities of LLMs?}
To evaluate the planning capabilities and transferability of LLMs precisely, we prepared multiple testing tasks with varying numbers of blocks in Blocksworld. Specifically, there are 100, 500, 500, and 500 testing tasks for 3-block, 4-block, 5-block, and 6-block settings, respectively. Meanwhile, the Logistics domain has too many variables (e.g., the number of cities, locations, trucks, airplanes, etc.) that affect task complexity, making it challenging to identify which variable impacts task difficulty the most. Therefore, we randomly selected 300 instances with varying numbers of variables as the testing data for the Logistics domain.

\subsubsection{Significant Improvement Achieved Through Fine-tuning}
We first assessed the planning capabilities of LLMs prior to fine-tuning, with results shown in Table \ref{table:before_after_finetuning}. Notably, LLMs without fine-tuning exhibit weak planning capabilities, and these results align with existing criticisms of LLMs' planning capabilities in the literature. 

Next, we evaluated the planning capabilities of LLMs after fine-tuning them on planning tasks using two sample sizes: small (100 samples) and large (1000 samples). Table \ref{table:before_after_finetuning} presents the test performance of LLMs after fine-tuning. Remarkably, even with a small number of samples, fine-tuning significantly enhanced the LLMs' planning capabilities. Additionally, we tested the fine-tuned models on the dataset from ~\cite{valmeekam2024planning}, where \texttt{GPT-3.5-turbo} and \texttt{Llama-3-8b} achieved solved rates of 90.6\% and 84.3\%, respectively. Humans are reported in ~\cite{valmeekam2024planning} to have a solved rate of 78\%. These findings challenge the prevailing notion that LLMs are inherently weak at planning. 

After closely investigating the failure cases of pre-fine-tuning models, we found that the failures were not due to poor instruction following or template mismatches, but because the LLMs could not provide correct plans. In contrast, fine-tuning even on a small amount of data significantly unlocks LLMs' planning capabilities. This suggests that LLMs' poor initial performance is due to a lack of exposure to automated planning tasks during pre-training. 

We believe that high-level planning skills cannot be acquired through prompting techniques or pipeline designs alone. Models fine-tuned on planning domains can serve as the backbone for developing more advanced applications. This strategy is more effective than solely relying on prompting strategies or pipelines. This critical insight, often overlooked in the community, provides valuable guidance on how to effectively enhance LLMs' planning capabilities.

\subsubsection{Data Scaling Effect}
The empirical results of the data scaling effect in both Blocksworld and Logistics are provided in Appendix \ref{app:data_scaling} due to space limits. Overall, the scaling effect is similar to what we have shown in the motivation example. LLM's planning capabilities improve as the amount of fine-tuning data increases. This is because the responses in fine-tuning data are of optimal quality, regarding accuracy, efficiency, and validity. However, this improvement follows an asymptotic pattern -- while initial increases in sample size lead to substantial performance gains, the rate of improvement (slope) diminishes as more samples are added. This suggests that exponentially increasing amounts of data are needed to further enhance LLMs' planning capabilities.

\subsubsection{Transferability}
We also examined the transfer performance of LLMs across different tasks and domains after fine-tuning. Due to space constraints, detailed cross-domain transferability results are presented in Appendix \ref{app:cross_domain_transfer}.

\textbf{In-domain Transferability:} We assessed LLM's transfer performance to different tasks within the same domain after fine-tuning. Specifically, \texttt{GPT-3.5-turbo} and \texttt{Llama-3-8b} were fine-tuned with 1,000 samples of 4-block, 5-block, and 6-block tasks, and then assessed on tasks with different block numbers than those used in fine-tuning. Note that we did not fine-tune the base models with 1,000 3-block samples because the 3-block setting does not provide enough distinct samples. The in-domain transferability results are shown in Table \ref{table:transferability_blocksworld}, and we make two key observations:
\begin{enumerate}
    \item LLMs exhibit good in-domain transferability and can apply their acquired planning capabilities to new challenges. For example, GPT-3.5 fine-tuned with 5-block tasks transfers well (68.0\%) to 6-block tasks, even when the tasks involve unseen block colors.  
    \item LLMs show better ``strong-to-weak" transferability than ``weak-to-strong" transferability. Specifically, LLMs transfer more effectively to less complex tasks. For instance, both GPT-3.5-6block and Llama-3-6block demonstrate better overall performance on tasks. This suggests that fine-tuning LLMs on more complex planning tasks can facilitate better transfer to simpler tasks.
\end{enumerate}

\begin{table}[th]
\setlength\tabcolsep{1.5pt} 
  \centering
  \begin{tabular}{cllll}
    \toprule
    \textbf{Model} &\textbf{3-block} &\textbf{4-block} &\textbf{5-block} &\textbf{6-block} \\
    \midrule
    GPT-4block       &98.0\% &91.9\% &65.8\% &34.0\% \\
    GPT-5block	     &81.0\% &88.0\% &84.8\% &68.0\% \\
    GPT-6block       &74.0\% &75.4\% &77.0\% &70.6\% \\
    \midrule              
    Llama-4block       &85.0\% &91.3\% &46.4\% &15.0\% \\
    Llama-5block	   &73.0\% &82.3\% &71.6\% &39.2\% \\
    Llama-6block       &63.0\% &65.6\% &69.2\% &58.6\% \\
    \bottomrule
  \end{tabular}
  \caption{In-domain transferability of fine-tuned LLMs. Models fine-tuned on $x$-block tasks are denoted as GPT-$x$block and Llama-$x$block. Fine-tuned models are evaluated on $x$-block tasks within the Blocksworld domain.}
  \label{table:transferability_blocksworld}
\end{table}

\begin{figure*}[ht]
  \centering
  \includegraphics[width=1.0\linewidth]{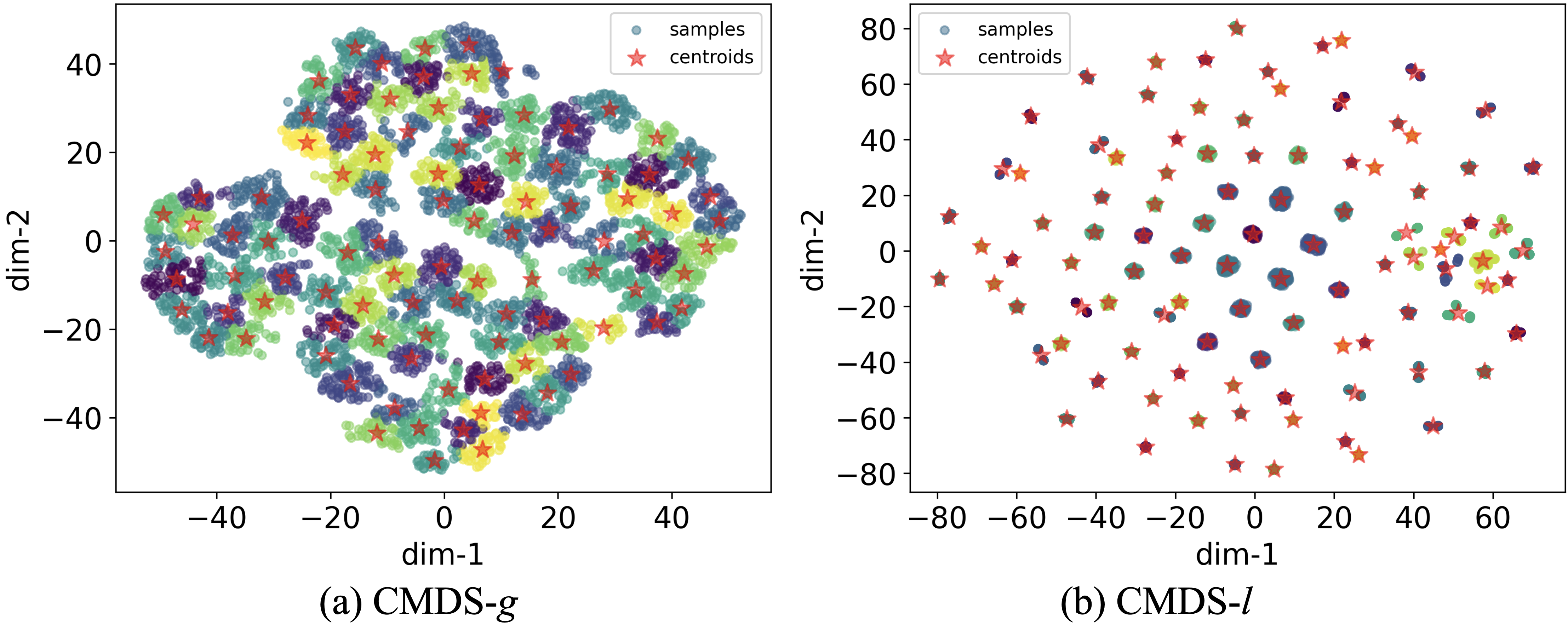}
  \caption{Subset samples selected by (a) CMDS-$g$ and (b) CMDS-$l$. Each point represents a planning task in Blocksworld, encoded in the embedding space. The sample point closest to each cluster centroid is selected for the subset. Note that clusters in CMDS-$l$ appear to have only a few dots as the points within each cluster are so closely packed together.}
  \label{fig:kmeans_clustering_blocksworld}
\end{figure*}

\textbf{Cross-domain Transferability:} While LLMs exhibit high in-domain transferability, our findings indicate their cross-domain transferability remains limited. We provide extensive results on this in Appendix \ref{app:extended_experiments}.3. Acquiring generalizable capabilities across different automated planning domains is particularly challenging for LLMs due to the substantial differences in actions, predicates, and objects. To address this, we fine-tuned LLMs on tasks from multiple domains collectively, aiming to develop multi-task capabilities. A detailed analysis of how data composition influences fine-tuning outcomes is included in the extended experimental section in Appendix \ref{app:extended_experiments}.4. In summary, although fine-tuning on mixed data effectively equips LLMs with multi-task planning capabilities, it introduces slightly higher variance in performance outcomes. 

\begin{table*}[th]
  \centering
  \begin{tabular}{ccccccc}
    \toprule
    & & & \multicolumn{4}{c}{\textbf{Subset Size}}                   \\
    \cmidrule(r){4-7}
    \textbf{Domain} &\textbf{Base Model} &\textbf{Algorithm} &$k=100$ &$k=200$ &$k=400$ &$k=1000$ \\
    \midrule
    \multirow{9}{*}[-0.5em]{\textbf{Blocksworld}} 
    & \multirow{3}*{GPT-3.5-turbo} & Random &61.8$\pm$1.9\% &72.4$\pm$1.8\% &81.9$\pm$1.5\% &91.9$\pm$1.2\% \\
    & & CMDS-$l$ &60.4$\pm$1.6\% &73.2$\pm$1.4\% &82.8$\pm$1.3\% &92.5$\pm$1.0\% \\
    & & CMDS-$g$ &\textbf{71.7$\pm$1.7\%} &\textbf{77.6$\pm$2.0\%} &\textbf{85.8$\pm$1.3\%} &\textbf{94.9$\pm$0.8\%} \\
    \cmidrule(r){2-7}
    & \multirow{3}*{Llama-2-7b} & Random &28.8$\pm$3.6\% &30.3$\pm$8.6\% &50.0$\pm$1.0\% &74.6$\pm$3.7\% \\
    & & CMDS-$l$ &24.8$\pm$4.4\% &32.6$\pm$3.8\% &51.5$\pm$1.4\% &74.2$\pm$2.9\% \\
    & & CMDS-$g$ &\textbf{32.5$\pm$2.1\%} &\textbf{37.2$\pm$4.5\%} &\textbf{52.9$\pm$1.3\%} &\textbf{75.7$\pm$0.7\%} \\
    \cmidrule(r){2-7}
    & \multirow{3}*{Llama-3-8b} & Random &43.9$\pm$2.9\% &54.2$\pm$6.8\% &67.6$\pm$3.3\% &86.3$\pm$0.8\% \\
    & & CMDS-$l$ &42.0$\pm$3.9\% &53.6$\pm$5.4\% &69.2$\pm$4.5\% &87.1$\pm$0.7\% \\
    & & CMDS-$g$ &\textbf{51.2$\pm$5.3\%} &\textbf{59.5$\pm$3.7\%} &\textbf{71.6$\pm$2.4\%} &\textbf{88.5$\pm$0.7\%} \\
    
    \midrule
    
    \multirow{9}{*}[-0.5em]{\textbf{Logistics}} 
    & \multirow{3}*{GPT-3.5-turbo} & Random &7.5$\pm$2.5\% &20.6$\pm$2.1\% &45.8$\pm$3.8\% &62.3$\pm$2.3\% \\
    & & CMDS-$l$ &8.3$\pm$1.5\% &19.3$\pm$2.7\% &42.2$\pm$2.1\% &66.3$\pm$1.7\% \\
    & & CMDS-$g$ &\textbf{11.1$\pm$1.6\%} &\textbf{25.0$\pm$1.2\%} &\textbf{51.8$\pm$1.2\%} &\textbf{73.0$\pm$1.1\%} \\
    \cmidrule(r){2-7}
    & \multirow{3}*{Llama-2-7b} & Random &0.4$\pm$0.1\% &1.9$\pm$0.3\% &11.9$\pm$1.3\% &53.0$\pm$2.1\% \\
    & & CMDS-$l$ &\textbf{0.8$\pm$0.3\%} &2.0$\pm$0.4\% &12.6$\pm$3.0\% &54.1$\pm$2.4\% \\
    & & CMDS-$g$ &0.4$\pm$0.2\% &\textbf{4.7$\pm$0.9\%} &\textbf{15.5$\pm$1.5\%} &\textbf{56.1$\pm$1.8\%} \\
    \cmidrule(r){2-7}
    & \multirow{3}*{Llama-3-8b} & Random &2.8$\pm$1.1\% &10.3$\pm$0.7\% &44.4$\pm$1.5\% &72.7$\pm$3.1\% \\
    & & CMDS-$l$ &3.9$\pm$1.0\% &8.8$\pm$1.5\% &46.8$\pm$3.8\% &73.2$\pm$1.8\% \\
    & & CMDS-$g$ &\textbf{5.3$\pm$1.0\%} &\textbf{12.9$\pm$1.6\%} &\textbf{49.8$\pm$2.6\%} &\textbf{75.3$\pm$0.8\%} \\
    \bottomrule
  \end{tabular}
  \caption{Task solved rate in the Blocksworld and Logistics domain. Results are presented as mean $\pm$ standard deviation. GPT and Llama results are collected across three and five random seeds respectively.}
  \label{table:algo_comparison_blocksworld_logistics}
\end{table*}

\subsection{RQ-2: Can we identify the most effective samples to enhance sample efficiency?}
In this section, we demonstrate that CMDS-$g$ identifies the most representative and effective samples, thereby enhancing sample efficiency during fine-tuning and improving the model's generalization capability afterward. Additionally, we found that CMDS-$l$ also outperforms random sampling, highlighting the value of the CMDS approach. We simplified testing data composition for Blocksworld, as detailed performance on different types of tasks is unnecessary for analysis in this section. We uniformly sampled from the four types of testing tasks (i.e., 3-block, 4-block, 5-block, 6-block) to collectively create a testing dataset with 1000 instances. Testing data for Logistics remains consistent with the previous sections.

First, we illustrate the effectiveness of the clustering-based method in solving the maximum diversity problem. Figure \ref{fig:kmeans_clustering_blocksworld} shows the subsets of samples identified by CMDS-$g$ and CMDS-$l$. In both plots, the selected samples (red stars) are located nearest to the centroids of each cluster, ensuring a diverse and broad coverage of the entire vector space. Notably, the data points in each cluster for CMDS-$l$ (Figure \ref{fig:kmeans_clustering_blocksworld}b) are very dense and closely packed, reflecting the limitations of language embeddings in capturing the complexity and structure of planning tasks. This results in clusters that are more influenced by narrative differences than by task structures, leading to a lack of diversity in the selected samples regarding task structure. In contrast, CMDS-$g$ (Figure \ref{fig:kmeans_clustering_blocksworld}a) utilizes graph embeddings to represent tasks with greater expressiveness. This approach allows CMDS-$g$ to capture the structural nuances of planning tasks more effectively, resulting in more representative clustering and a more diverse selection of samples.

Next, we empirically demonstrate the effectiveness of CMDS-$g$ at different scales. In Blocksworld, we created five thousand instances with varying numbers of blocks as the training dataset. Using Random, CMDS-$l$, and CMDS-$g$, we selected task subsets of varying sizes ($k=$100, 200, 400, and 1,000), fine-tuned the base models, and then assessed their performance on the hold-out testing data. The training and testing data for Logistics are the same as in previous experiments. Comprehensive results for both Blocksworld and Logistics are provided in Tables \ref{table:algo_comparison_blocksworld_logistics}. 

Notably, CMDS-$g$ consistently outperforms both Random and CMDS-$l$ across different scales in both domains. CMDS-$g$ significantly improves sample efficiency in fine-tuning, particularly at smaller scales. For instance, CMDS-$g$ with $k=100$ achieves a 71.7\% solved rate, a performance level that requires approximately 200 samples when using Random or CMDS-$l$. While the performance advantage of CMDS-$g$ diminishes with larger sample sizes—reflecting the diminishing returns from additional samples—this is attributed to the nature of fine-tuning rather than a limitation of CMDS-$g$. Therefore, CMDS-$g$ is particularly advantageous in scenarios where fine-tuning incurs high economic and computational costs, making it ideal for applications with limited resources or strict efficiency requirements. Furthermore, as highlighted in the motivation example, CMDS-$g$ remains valuable as task complexity increases, requiring significantly less data to converge and generalize, thereby reducing costs substantially.

\begin{figure}[ht]
  \centering
  \includegraphics[width=1.0\linewidth]{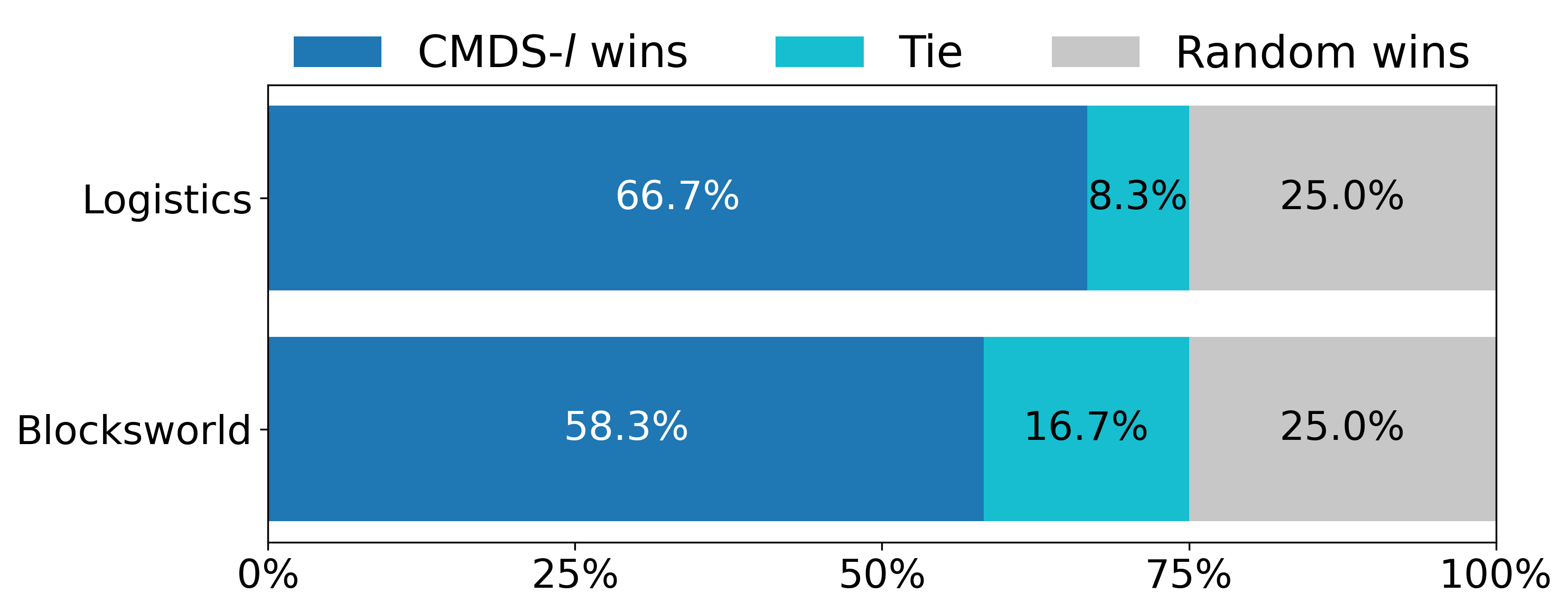}
  \caption{Performance comparison between CMDS-$l$ and Random.}
  \label{fig:win_tie_los}
\end{figure}

Finally, we present a detailed comparison of the performance between CMDS-$l$ and Random. Figure \ref{fig:win_tie_los} illustrates the win-tie-loss comparison between these methods in the Blocksworld and Logistics domains. CMDS-$g$ is excluded from this comparison because it consistently outperforms both. The results indicate that CMDS-$l$ achieves more wins across both domains, further validating the effectiveness of the CMDS approach. Additionally, due to its broad applicability of language embeddings, CMDS-$l$ is valuable for fine-tuning LLMs in domains beyond automated planning.

\section{Conclusion}
In this paper, we extensively study how data scaling, diversity, and task complexity affect fine-tuning outcomes in automated planning domains. Our findings show that LLMs can exhibit strong planning capabilities through fine-tuning with sufficiently large datasets, challenging the notion that LLMs are inherently weak in System 2 planning. To enhance sample efficiency and reduce fine-tuning costs, we introduce the Clustering-Based Maximum Diversity Sampling (CMDS) approach, which ensures a broad and representative sample set. Our methods, CMDS-$g$ (using graph embeddings) and CMDS-$l$ (using language embeddings), consistently outperform random sampling. Notably, CMDS-$g$ significantly enhances both the sample efficiency in fine-tuning and the generalization performance of the fine-tuned models. This research provides a thorough investigation of LLMs' planning capabilities and offers insights into effectively and efficiently developing these capabilities.

\section{Limitations}
While the proposed algorithm enhances LLMs' performance in automated planning with higher sample efficiency and better generalization performance, the planning capabilities acquired in one domain can hardly transfer to new domains. Although the mixed-data fine-tuning approach proves effective, it also introduces slightly higher variance in outcomes, indicating the need for further refinement and stability enhancements. Addressing these limitations will be the focus of our future work. Cross-domain transferability in planning remains an underexplored area, highlighting a significant opportunity for further investigation.


\section*{Acknowledgments}
This research/project is supported by the National Research Foundation Singapore and DSO National Laboratories under the AI Singapore Programme (AISG Award No: AISG2-RP-2020-017).

\bibliography{custom}

\appendix
\appendix
\onecolumn

\section{Related Work} \label{app:related_work}
\paragraph{Enhancing LLMs' Reasoning and Planning Capabilities with Prompting.} Prompting involves guiding the input to a Large Language Model (LLM) using templates or cues to produce the desired output. Techniques like Chain of Thought (CoT, ~\cite{wei2022chain}) and Tree of Thoughts (ToT, ~\cite{yao2024tree}) have demonstrated the ability to enhance complex reasoning and planning by breaking down tasks into intermediate reasoning steps. These prompting methods have proven effective in solving mathematical problems \cite{huang2022language,yao2022react} and household tasks \cite{wei2022chain,yao2024tree}. However, their effectiveness in automated planning domains has been limited, largely due to the inherent complexity of these tasks. For example, ~\cite{valmeekam2024planning} showed that CoT prompting resulted in only a 1\% improvement in Blocksworld, with GPT-4~\cite{achiam2023gpt} achieving a 34.6\% solved rate without CoT prompting and 35.6\% with it. 

\paragraph{LLMs with access to external tools.} Due to LLMs' tendency to generate responses with hallucinations and their lack of strict adherence to actions and predicates, some approaches integrate external tools to provide feedback. For example, ~\cite{liu2023llm+} and ~\cite{guan2023leveraging} utilize the Planning Domain Definition Language (PDDL) executor to verify the validity of generated plans and provide feedback (such as explanations for why plans are not executable) to aid LLMs in reflection and plan refinement. However, these methods do not fundamentally enhance the LLMs' inherent planning capabilities and result in complex integration engineering and potential high latency. Furthermore, frequent use of external tools can be costly and prone to reliability issues. For instance, the effectiveness of traditional planners depends heavily on the accuracy of translating natural language to PDDL; inaccuracies in this translation can lead to incorrect plans or failed validations.

\paragraph{Enhancing LLM's task-specific capabilities with fine-tuning.} Fine-tuning refers to the process of updating the parameters of a pre-trained model using task-specific data. Initial studies, such as Less Is More for Alignment (LIMA, ~\cite{zhou2024lima}), have investigated the effects of training data quantity and quality on fine-tuning outcomes in typical natural language processing tasks like Q\&A and creative writing. However, to the best of our knowledge, there has been no comprehensive study on how fine-tuning data (considering aspects like quantity, diversity, composition, etc.) affects the planning capabilities of LLMs. This paper pioneers the exploration of these research questions.

\paragraph{Large Reasoning Models.} 
Recently, OpenAI released o1-preview~\cite{openai_o1_preview}, a model trained to generate an internal thought before answering questions using effective reinforcement learning with human feedback (RLHF) techniques. o1 pioneers the application of scaling laws during inference and showcases a notable improvement in reasoning and planning capabilities compared to existing LLMs. Our proposed method focuses on enhancing the base model during the supervised fine-tuning (SFT) stage, a critical step in preparing a strong initial model for RLHF. Given the proven effectiveness of CMDS in the SFT stage, we believe that combining CMDS with advanced RLHF techniques could further enhance the model's reasoning and planning performance and potentially reduce overall training costs. We consider this integration a potential direction for future work.

\paragraph{Data Selection.} 
Existing work on data selection for training LLMs, such as ~\cite{yu2023cold} and ~\cite{zhu2022unsupervised}, addresses related challenges but differs from ours in two key ways. First, our work focuses specifically on enhancing logical reasoning and planning capabilities in LLMs, a critical and increasingly important domain. Our data selection method has the potential to be applied to training large reasoning models like o1, whereas ~\cite{yu2023cold} centers on text classification and ~\cite{zhu2022unsupervised} targets question generation from text passages. Second, and more importantly, our approach differs fundamentally in methodology. Both ~\cite{yu2023cold} and ~\cite{zhu2022unsupervised} rely on natural language embeddings, similar to our baseline method, CMDS-$l$. However, as demonstrated in Section 4.2, language embedding-based methods struggle to effectively distinguish nuanced planning instances, highlighting the limitations of these approaches in our targeted domain.

\section{Extended Experiment Results} \label{app:extended_experiments}
In this section, we present extended experimental results, covering various aspects of our proposed method and details about the planning capabilities of LLMs. These include its performance on imbalanced datasets (\ref{app:imbalanced_data}), the detailed effects of dataset scaling (\ref{app:data_scaling}), cross-domain transferability (\ref{app:cross_domain_transfer}), the impact of data composition on fine-tuning outcomes (\ref{app:data_composition}), an analysis of the optimality of plans generated by LLMs (\ref{app:plan_optimality}), a detailed performance analysis (\ref{app:detailed_performance}), further investigation into the limitations of language embeddings (\ref{app:language_embeddings}), and a visualized evaluation via attention (\ref{app:visualized_eval}).

\subsection{Imbalanced Dataset} \label{app:imbalanced_data}
\begin{figure}[ht]
  \centering
  \includegraphics[width=0.5\linewidth]{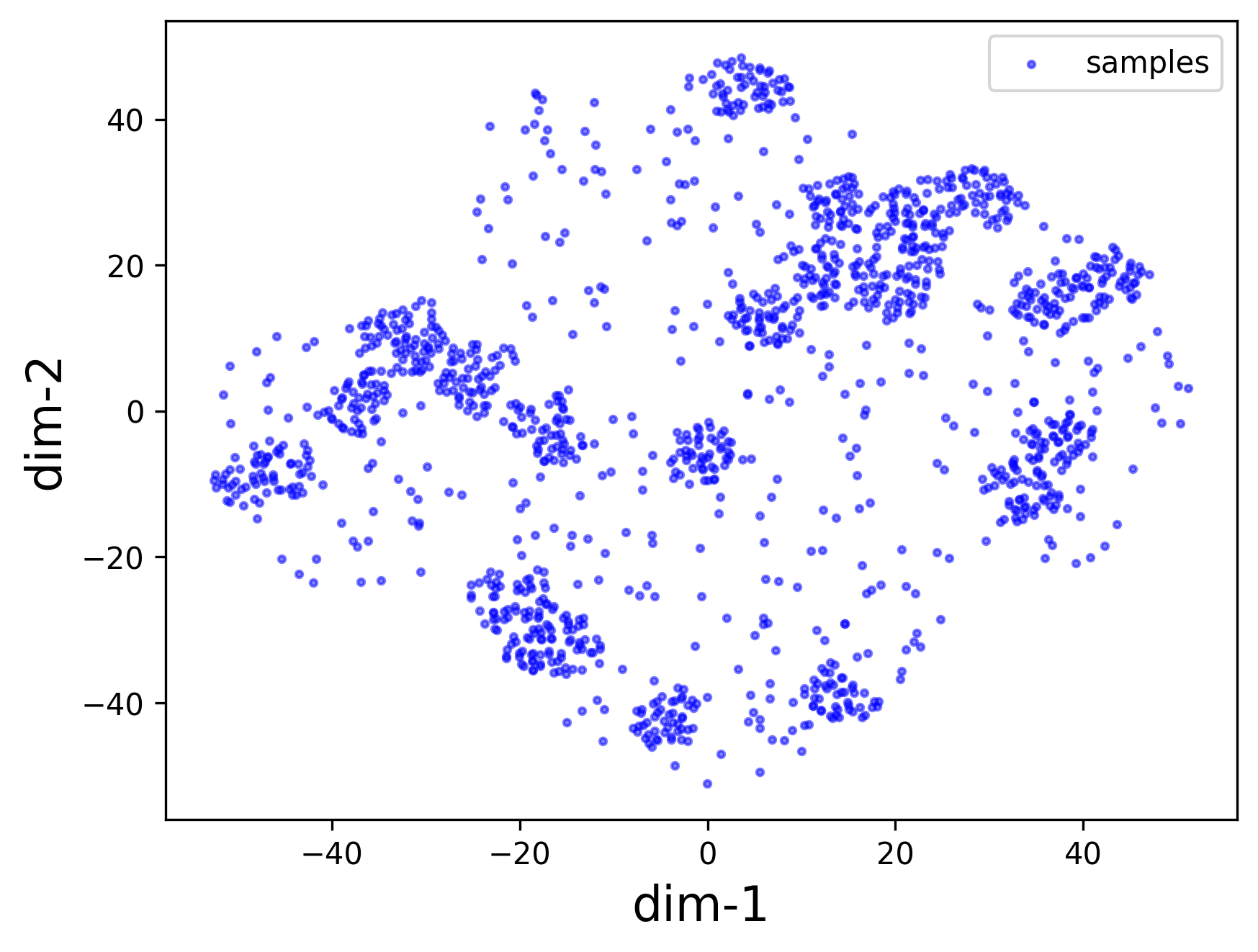}
  \caption{Imbalanced data distribution in the Blocksworld domain.}
  \label{fig:bw_biased_dataset}
\end{figure}

The fine-tuning data in the previous experiments were collected through randomization with unique checking, ensuring a uniform and balanced training dataset. However, in real-world scenarios, fine-tuning data often varies in quality and distribution. To simulate this, we conducted additional experiments using imbalanced fine-tuning data. Since no standard metric exists to measure dataset imbalance in automated planning, we devised a heuristic imbalance coefficient to control the level of imbalance. Specifically, we randomly removed data points from $p$*100\% of all clusters (identified by CMDS-$g$ such that the samples represent diverse planning instances based on task structures) to create datasets with imbalanced distributions. Within the clusters selected for data removal, we retained $j \in \left[1, 5\right]$ points. As shown in Figure \ref{fig:bw_biased_dataset}, this process resulted in areas with dense examples and others with sparse examples.

We created datasets with varying degrees of imbalance by adjusting the percentage of clusters removed from the training dataset. Note that this imbalance coefficient is an approximate metric designed to control the dataset's imbalance level, and it does not strictly scale linearly with the degree of imbalance.

\begin{figure}[ht]
  \centering
  \includegraphics[width=0.5\linewidth]{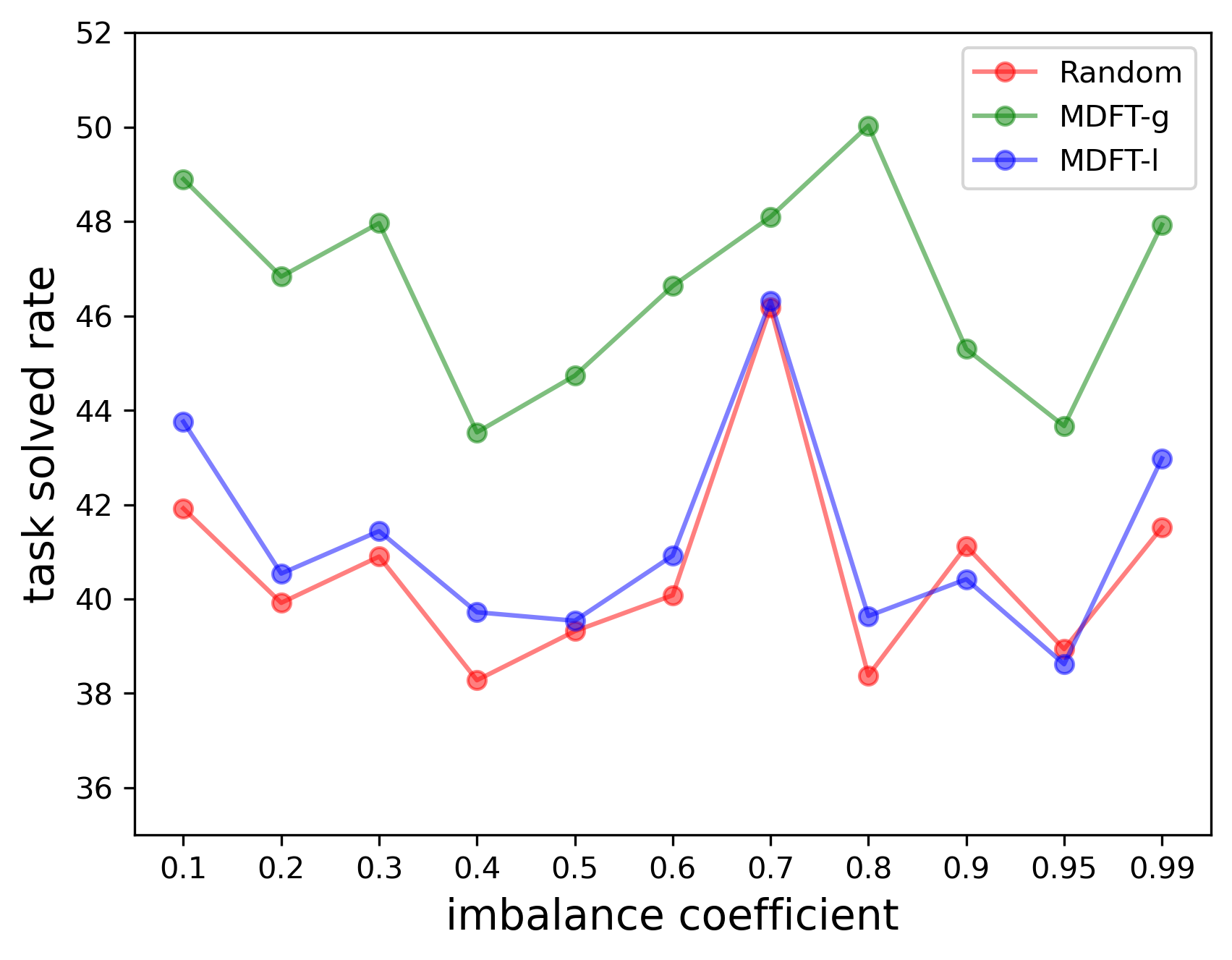}
  \caption{Fine-tuning performance vs imbalance coefficient when $k=100$ in Blocksworld. Results are collected with Llama-3-8b and averaged across five independent runs.}
  \label{fig:perf_bias_degree_bw}
\end{figure}

We employed Random, CMDS-$l$, and CMDS-$g$ sampling methods on the imbalanced datasets and conducted experiments using Llama-3-8b. The results, depicted in Figure \ref{fig:perf_bias_degree_bw}, lead to two key observations: (1) CMDS-$g$ consistently outperforms both Random and CMDS-$l$ by a significant margin; (2) generally, CMDS-$l$ outperforms Random, winning in 9 out of 11 imbalance coefficient scales. Additionally, the results shown in Figure \ref{fig:perf_bias_degree_bw} are worse compared to those obtained with a balanced dataset, indicating that a balanced dataset is beneficial for enhancing model capabilities.

\subsection{Data Scaling Effect} \label{app:data_scaling}
\begin{figure}[ht]
  \centering
  \includegraphics[width=0.9\linewidth]{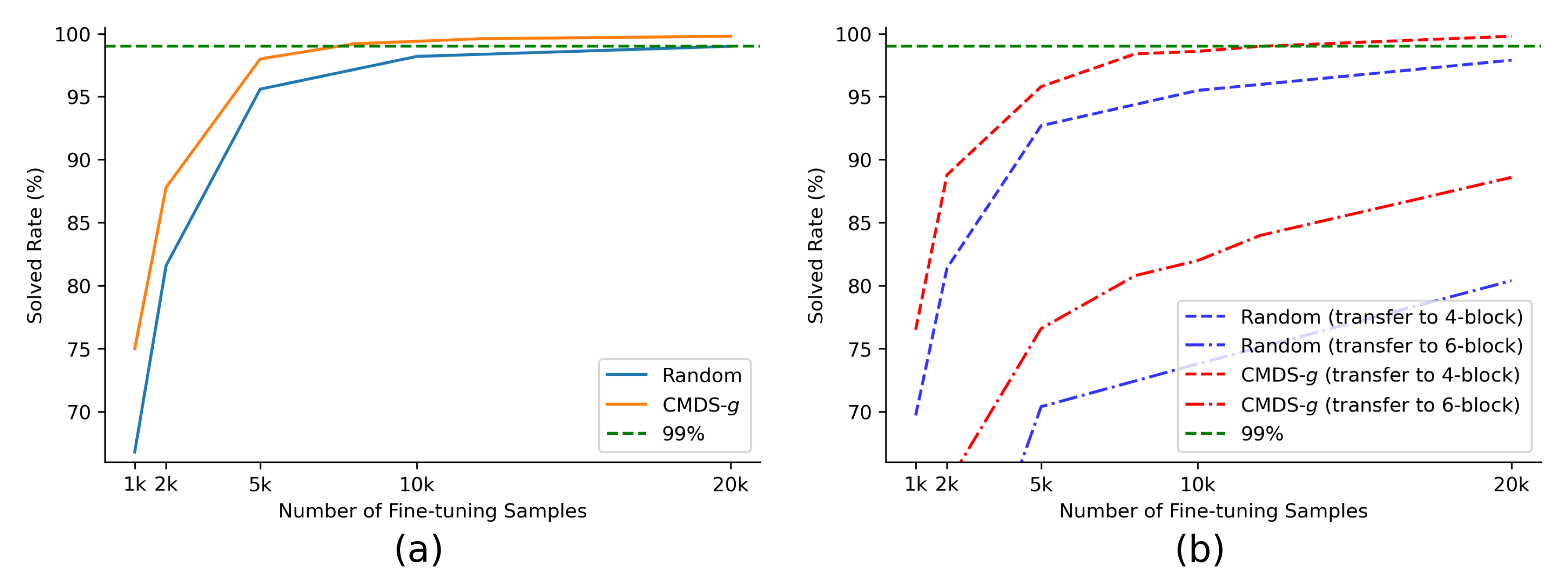}
  \caption{Scaling effect of fine-tuning Llama-3-8b in Blocksworld. (a) Model fine-tuned on 5-block samples and evaluated on hold-out 5-block test tasks. (b) Model fine-tuned on 5-block samples and evaluated on hold-out 4-block and 6-block test tasks.}
  \label{fig:performance_vs_size_transfer}
\end{figure}

In this section, we first assess the scaling effect on the model's in-domain transferability. Specifically, we evaluate models trained on 5-block tasks (as shown in the motivation example) using held-out 4-block and 6-block test tasks, with the results presented in Figure \ref{fig:performance_vs_size_transfer}b. We make three key observations: (1) Transfer performance follows the same scaling effect, exhibiting an asymptotic pattern—initial increases in sample size result in significant performance gains, but the rate of improvement diminishes as more samples are added; (2) Transfer performance is generally worse than standard performance, i.e., test performance on tasks that match the training settings; (3) CMDS-$g$ achieves higher sample efficiency in transfer performance. For instance, CMDS-$g$ attains a 99\% solved rate with approximately 7,500 samples, whereas Random requires around 20,000 samples to reach this performance. More importantly, using fine-tuning data selected by CMDS-$g$ enables the model to generalize better to different tasks. These advantages underscore the value of our proposed method, particularly as task complexity increases and the data requirements rise exponentially. Our approach significantly enhances sample efficiency and generalization capability while reducing the substantial time, economic, and computational costs involved.

\begin{figure}[ht]
  \centering
  \includegraphics[width=0.5\linewidth]{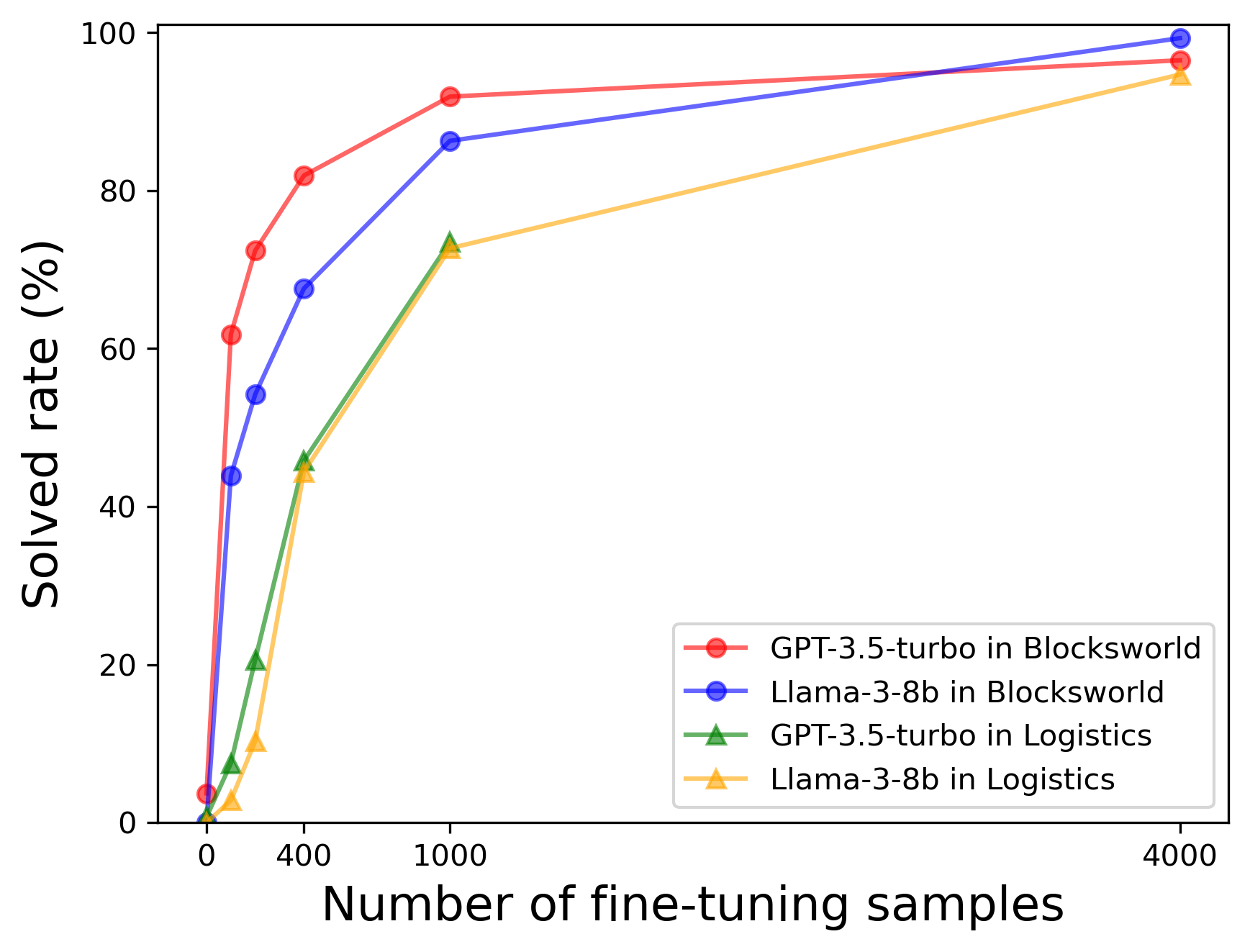}
  \caption{Model performance versus the number of fine-tuning samples. GPT-3.5-turbo finetuned with 4000 Logistics samples is omitted in this experiment due to high economic costs.}
  \label{fig:performance_vs_size}
\end{figure}

Furthermore, we closely examine the impact of data scaling on the planning capabilities of LLMs using a smaller-scale experiment. We conducted tests in the 4-blocks setting of Blocksworld with GPT-3.5-turbo-0125 and Llama-3-8b. The results are summarized in Figure \ref{fig:performance_vs_size}. Our findings indicate that GPT-3.5-turbo exhibits superior planning capabilities compared to the Llama models, both before and after fine-tuning. However, when fine-tuned with 4000 samples, Llama-3-8b outperforms GPT-3.5-turbo. This discrepancy is largely due to the limited number of fine-tuning epochs applied to GPT-3.5-turbo, driven by the high economic costs, while the Llama models were fine-tuned over significantly more epochs.

As observed consistently, LLMs' planning capabilities improve with increased fine-tuning data, due to the high quality of responses in terms of accuracy, efficiency, and validity. However, this improvement follows an asymptotic pattern, where the rate of performance gains diminishes as more samples are added. This indicates that substantial data is needed to fine-tune LLMs effectively for complex System 2 planning tasks, with data requirements growing exponentially with task complexity. In such cases, our proposed method can significantly reduce the time, economic, and computational costs associated with fine-tuning.


\subsection{Cross-domain Transferability} \label{app:cross_domain_transfer}
To evaluate cross-domain transferability, we fine-tuned the base models in one domain and then tested their performance in another. Specifically, we fine-tuned the models on 200 tasks from the Blocksworld domain and evaluated their performance in the Logistics domain. Conversely, we fine-tuned the models on 400 tasks from the Logistics domain and assessed their performance in the Blocksworld domain. We intentionally used relatively small sample sizes in this experiment to avoid overfitting the LLMs to the specific response patterns and dynamics of a single domain. The results of these cross-domain transferability tests are presented in Table \ref{table:transferability_across_domains}.

\begin{table*}[th]
  \centering
  \begin{tabular}{cllll}
    \toprule
    & & & \multicolumn{2}{c}{\textbf{Fine-tuning Tasks}}                   \\
    \cmidrule(r){4-5}
    \textbf{Test Domain}      & \textbf{Model} &\textbf{un-finetuned} &\textbf{Blocksworld} &\textbf{Logistics} \\
    \midrule
    \multirow{2}*{Blocksworld} &GPT-3.5 &3.7\% &72.4\% &3.1\% \\
		                      &Llama-3 &0\% &54.2\% &0.3\% \\
    \midrule
    \multirow{2}*{Logistics} &GPT-3.5 &0.7\% &0.3\% &45.8\% \\
		                &Llama-3 &0\% &0\% &44.4\% \\
    \bottomrule
  \end{tabular}
  \caption{LLM's transferability across domains.}
  \label{table:transferability_across_domains}
\end{table*}

The findings reveal that LLMs exhibit poor transferability across different planning domains after fine-tuning, in some cases performing worse than models that were not fine-tuned. This is due to the substantial differences between automated planning domains in terms of actions, predicates, and objects, which hinder the LLMs' ability to generalize across domains. To develop LLMs with effective multi-task capabilities, it is therefore necessary to fine-tune them on tasks drawn from multiple domains simultaneously. In the next section, we provide a detailed analysis of how data composition influences fine-tuning outcomes.

\subsection{Data Composition} \label{app:data_composition}
To assess the impact of data composition on fine-tuning, we created a mixed dataset by randomly sampling task instances from both the Blocksworld and Logistics domains. This mixed dataset was used to fine-tune GPT-3.5-turbo and Llama-3-8b models. The dataset was composed of an equal number of samples from each domain, which were randomly shuffled before fine-tuning. For example, a dataset with $k=100$ includes 100 samples from Blocksworld and 100 samples from Logistics, making it directly comparable to a dataset of $k=100$ pure samples from a single domain. The effects of data composition on fine-tuning outcomes are summarized in Table \ref{table:data_composition_blocksworld_logistics}. 

Our key finding is that fine-tuning on mixed-domain data does not degrade model performance; in fact, it slightly improves it. Statistically, mixed data outperforms pure data in 5 out of 7 settings in both the Blocksworld and Logistics domains. This result is significant because, despite the poor cross-domain transferability of LLMs, we can enhance their planning capabilities across different domains by fine-tuning on a combined dataset. This conclusion is both crucial and promising, indicating that LLMs can acquire strong cross-domain planning capabilities through fine-tuning on sufficiently large and diverse datasets from multiple domains. These findings underscore the potential of using mixed planning tasks to fine-tune LLMs for multi-task proficiency.

\begin{table*}[th]
  \centering
  \begin{tabular}{ccccccc}
    \toprule
    & & & \multicolumn{4}{c}{\textbf{Subset Size}}                   \\
    \cmidrule(r){4-7}
    \textbf{Domain} &\textbf{Model} &\textbf{Composition} &$k=100$ &$k=200$ &$k=400$ &$k=1000$ \\
    \midrule
    \multirow{4}{*}[-0.5em]{\textbf{Blocksworld}} 
    & \multirow{2}*{GPT} & pure data &\textbf{61.8$\pm$1.9\%} &72.4$\pm$1.8\% &81.9$\pm$1.5\% &91.9$\pm$1.2\% \\
                                 & & mixed data &58.8$\pm$1.6\% &\textbf{75.4$\pm$0.6\%} &\textbf{85.0$\pm$0.5\%} &-- \\
    \cmidrule(r){2-7}
    & \multirow{2}*{Llama} & pure data &\textbf{43.9$\pm$2.9\%} &54.2$\pm$6.8\% &67.6$\pm$3.3\% &86.3$\pm$0.8\% \\
                              & & mixed data &35.1$\pm$11.6\% &\textbf{58.7$\pm$4.8\%} &\textbf{79.6$\pm$3.3\%} &\textbf{89.6$\pm$1.5\%} \\
    \midrule
    \multirow{4}{*}[-0.5em]{\textbf{Logistics}} 
    & \multirow{2}*{GPT} & pure data &7.5$\pm$2.5\% &20.6$\pm$2.1\% &\textbf{45.8$\pm$3.8\%} & 62.3$\pm$2.3\% \\
    & & mixed data &\textbf{10.4$\pm$0.1\%} &\textbf{22.1$\pm$3.5\%} &36.4$\pm$2.8\% &-- \\
    \cmidrule(r){2-7}
    & \multirow{2}*{Llama} & pure data &0.4$\pm$0.1\% &10.3$\pm$0.7\% &45.4$\pm$0.5\% &\textbf{72.7$\pm$3.1\%} \\
    & & mixed data &\textbf{5.9$\pm$1.4\%} &\textbf{19.1$\pm$1.6\%} &\textbf{54.0$\pm$6.4\%} &68.0$\pm$5.1\% \\
    \bottomrule
  \end{tabular}
  \caption{Task solved rate in the Blocksworld and Logistics domain. $k$ is the number of fine-tuning samples selected from one domain. E.g., $k=100$ indicates 100 samples from Blocksworld and 100 samples from Logistics. Results are presented as mean $\pm$ standard deviation. GPT and Llama represent \texttt{GPT-3.5-turbo} and \texttt{Llama-3-8b} and their results are collected across three and five random seeds respectively. Results for GPT-3.5-turbo with $k=1000$ are omitted in this experiment due to the high economic cost.}
  \label{table:data_composition_blocksworld_logistics}
\end{table*}

Combining these findings with those discussed in Section A.2 on the Data Scaling Effect, we conclude that LLMs can achieve robust System 2 planning capabilities by being fine-tuned on sufficiently large datasets composed of diverse samples from various planning domains. Our research offers valuable insights to both the research community and industry on how to effectively and efficiently train LLMs to develop strong planning capabilities.

\subsection{Plan Optimality} \label{app:plan_optimality}
We evaluated the optimality of the plans generated by LLMs and found that they have a high likelihood of being optimal. As shown in Table \ref{table:optimality_rate}, GPT-3.5-turbo and Llama-3-8b achieved near-optimal performance when fine-tuned with 4,000 samples in both the Blocksworld and Logistics domains. For comparison, a human baseline in the Blocksworld domain, as reported by \cite{valmeekam2024planning}, showed that 50 human participants had a task success rate of 78.0\% and a plan optimality rate of 89.7\%. Given the higher complexity of the Logistics domain, the human baseline there would likely be lower than in Blocksworld. Remarkably, LLMs maintained high optimality rates—over 90\% in Blocksworld and over 80\% in Logistics—even when fine-tuned with just 100 samples. This result is notable and particularly relevant in the planning domain, where the optimal plan, often being the shortest, is likely to be the easiest for LLMs to identify. Achieving such superior optimality rates is challenging through prompting techniques or system designs alone, further emphasizing the importance of extensive fine-tuning. These results underscore the strong potential of LLMs in achieving System 2 planning capabilities.

\begin{table*}[th]
  \centering
  \begin{tabular}{cccccc}
    \toprule
    & & & \multicolumn{3}{c}{\textbf{Subset Size}}                   \\
    \cmidrule(r){4-6}
    \textbf{Domain} &\textbf{Base Model} &\textbf{Data Composition} &$k=100$ &$k=400$ &$k=4000$ \\
    \midrule
    \multirow{4}{*}[-0.5em]{\textbf{Blocksworld}} 
    & \multirow{2}*{GPT-3.5-turbo} & Solved Rate &61.8\% &81.9\% &96.5\% \\
                                 & & Optimality Rate &89.3\% &92.4\% &98.4\% \\
    \cmidrule(r){2-6}
    & \multirow{2}*{Llama-3-8b} & Solved Rate &43.9\% &67.6\% &99.3\% \\
                              & & Optimality Rate &90.0\% &92.6\% &100\% \\
    \midrule
    \multirow{4}{*}[-0.5em]{\textbf{Logistics}} 
    & \multirow{2}*{GPT-3.5-turbo} & Solved Rate &7.5\% &45.8\% &-- \\
                                 & & Optimality Rate &78.3\% &98.1\% &-- \\
    \cmidrule(r){2-6}
    & \multirow{2}*{Llama-3-8b} & Solved Rate &43.9\% &67.6\% &94.7\% \\
                              & & Optimality Rate &81.2\% &91.2\% &99.3\% \\
    \bottomrule
  \end{tabular}
  \caption{Analysis of the optimality rate of generated plans in Blocksworld and Logistics domain. The samples are all randomly sampled in this experiment. The Optimality rate is calculated based on the correct plans, optimality rate = (number of optimal plans)/(number of correct plans). GPT and Llama results are averaged across three and five random seeds. Results for GPT-3.5-turbo with $k=4000$ in Logistics are omitted in this experiment due to the high economic cost.}
  \label{table:optimality_rate}
\end{table*}

\subsection{Detailed Performance} \label{app:detailed_performance}
\begin{figure}[ht]
  \centering
  \includegraphics[width=0.8\linewidth]{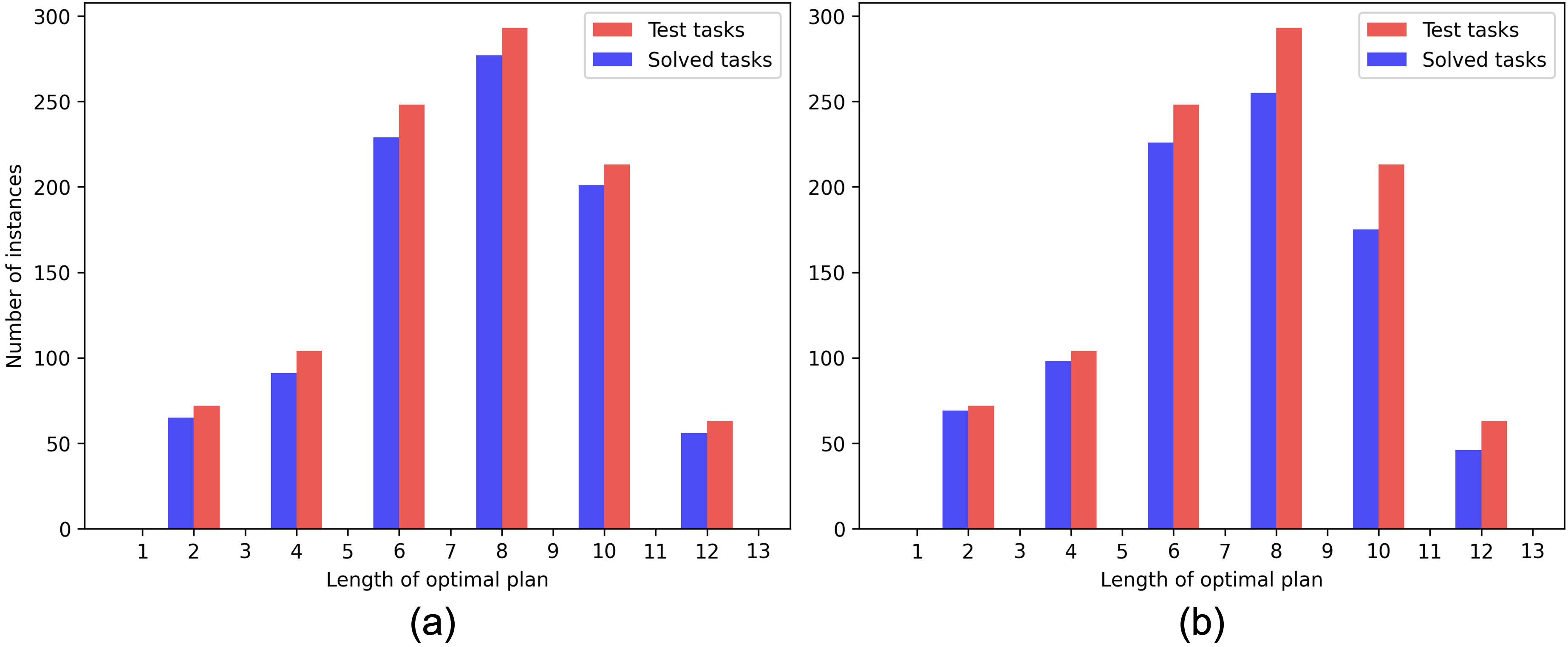}
  \caption{Distribution of the solved tasks in the Blocksworld domain. (a) GPT-3.5-turbo and (b) Llama-3-8b fine-tuned on 1000 4-block tasks and tested on 4-block tasks.}
  \label{fig:solved_dist_bw_1k}
\end{figure}

\begin{figure}[ht]
  \centering
  \includegraphics[width=0.8\linewidth]{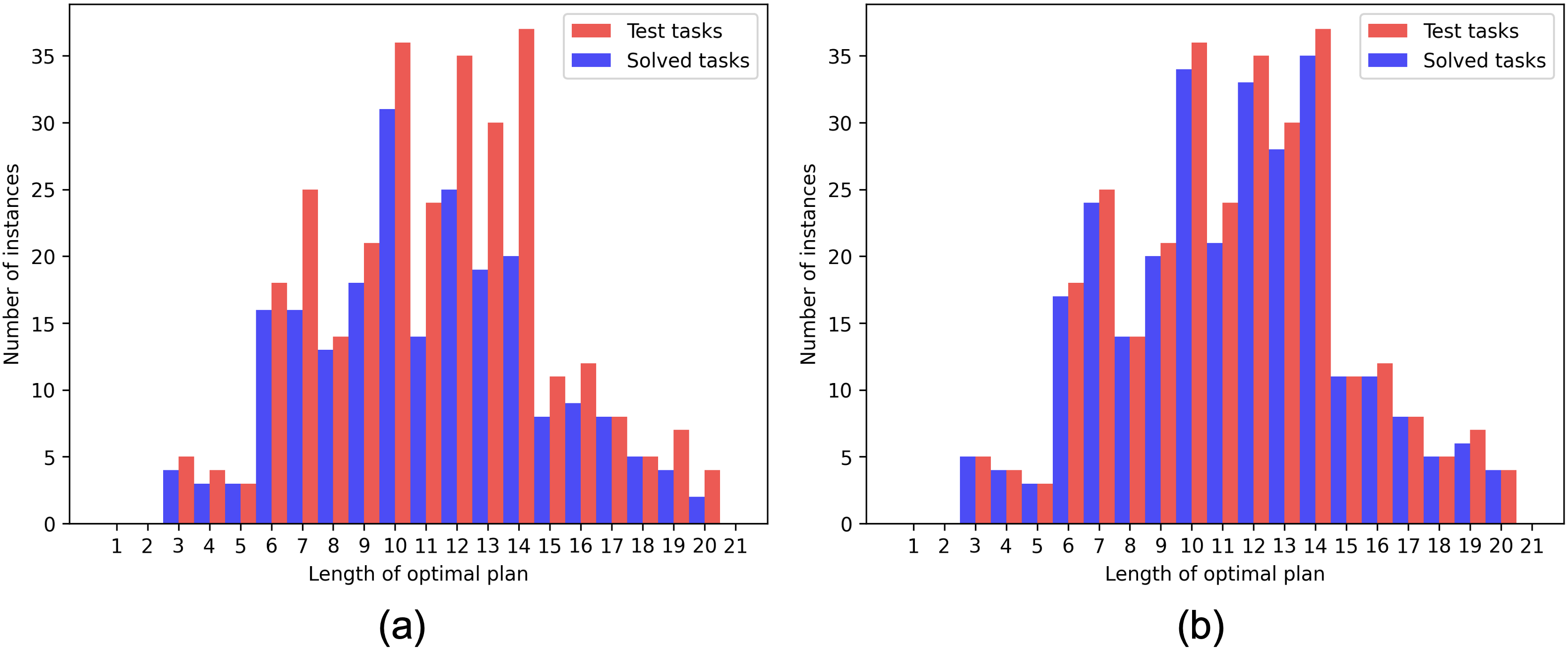}
  \caption{Distribution of the solved tasks in the Logistics domain. (a) GPT-3.5-turbo is fine-tuned on 1000 tasks, and (b) Llama-3-8b is fine-tuned on 4000 tasks.}
  \label{fig:solved_dist_log}
\end{figure}

Figure \ref{fig:solved_dist_bw_1k} and Figure \ref{fig:solved_dist_log} provide a detailed analysis of the LLMs' planning capabilities in the Blocksworld and Logistics domains. 

From Figure \ref{fig:solved_dist_bw_1k}, we observe that GPT's performance, as measured by the solved rate, remains consistent even as the optimal plan length increases, demonstrating robust and reliable planning capabilities. In contrast, Llama shows diminished performance on more complex tasks. 

Figure \ref{fig:solved_dist_log} further reveals that LLMs can have good planning capabilities across tasks of varying complexity (i.e., they can effectively solve both short and long planning tasks). GPT's performance is worse than Llama's because it is fine-tuned on less amount of data. We do not observe a clear pattern indicating that LLMs perform better on simpler tasks (i.e., those requiring fewer steps) than on more complex tasks (i.e., those requiring more steps). These findings suggest that once LLMs are fine-tuned on sufficiently large datasets, they can perform consistently well across tasks of varying complexity in terms of plan lengths. 

\subsection{Language Embeddings} \label{app:language_embeddings}
To generate language embeddings for planning tasks, we simplify the prompts by removing redundant information and only converting the initial and goal configurations into the embedding space to calculate the distance between tasks. Domain instructions are omitted since they are identical for all tasks. In our experiments, we used the RoBERTa~\cite{liu2019roberta} to transform natural language descriptions into the language embedding space.

Below is an example task in the Blocksworld domain and its altered version. The altered task slightly differs from the original task, with only the positions of the blue and yellow blocks switched. The L2 norm distance between the embeddings of these two tasks is 3.6e-07, which is too small to be effective in the downstream clustering process. This results in points within a cluster being closely packed, making it difficult to identify multiple representative points from a single cluster. This example highlights the phenomenon illustrated in Figure \ref{fig:kmeans_clustering_blocksworld}, where planning tasks converted into the language embedding space are densely packed, with clusters far apart from each other. This occurs because language embeddings are influenced more by narrative differences than by task structure.

Despite this shortcoming, CMDS-$l$ still outperforms the Random selection, indicating the effectiveness of the Clustering-Based Maximum Diversity Sampling. Our results further validate that diversity plays an important role in automated planning, as it does in many other tasks involving the fine-tuning of LLMs.

\begin{tcolorbox}[colback=gray!10, colframe=black, title=An Example Task in the Blocksworld]
\begin{tcolorbox}[colback=yellow!20, colframe=black, title=Original Task]
As initial conditions I have that, the red block is clear, the orange block is clear, the yellow block is clear, the hand is empty, the yellow block is on top of the blue block, the red block is on the table, the blue block is on the table and the orange block is on the table.
My goal is to have that the red block is on top of the orange block and the orange block is on top of the yellow block.
\end{tcolorbox}

\begin{tcolorbox}[colback=green!20, colframe=black, title=Altered Task]
As initial conditions I have that, the red block is clear, the orange block is clear, the blue block is clear, the hand is empty, the blue block is on top of the yellow block, the red block is on the table, the yellow block is on the table and the orange block is on the table.
My goal is to have that the red block is on top of the orange block and the orange block is on top of the yellow block.
\end{tcolorbox}
\end{tcolorbox}

\subsection{Visualized Evaluations via Attention} \label{app:visualized_eval}
\begin{figure}[ht]
  \centering
  \includegraphics[width=0.8\linewidth]{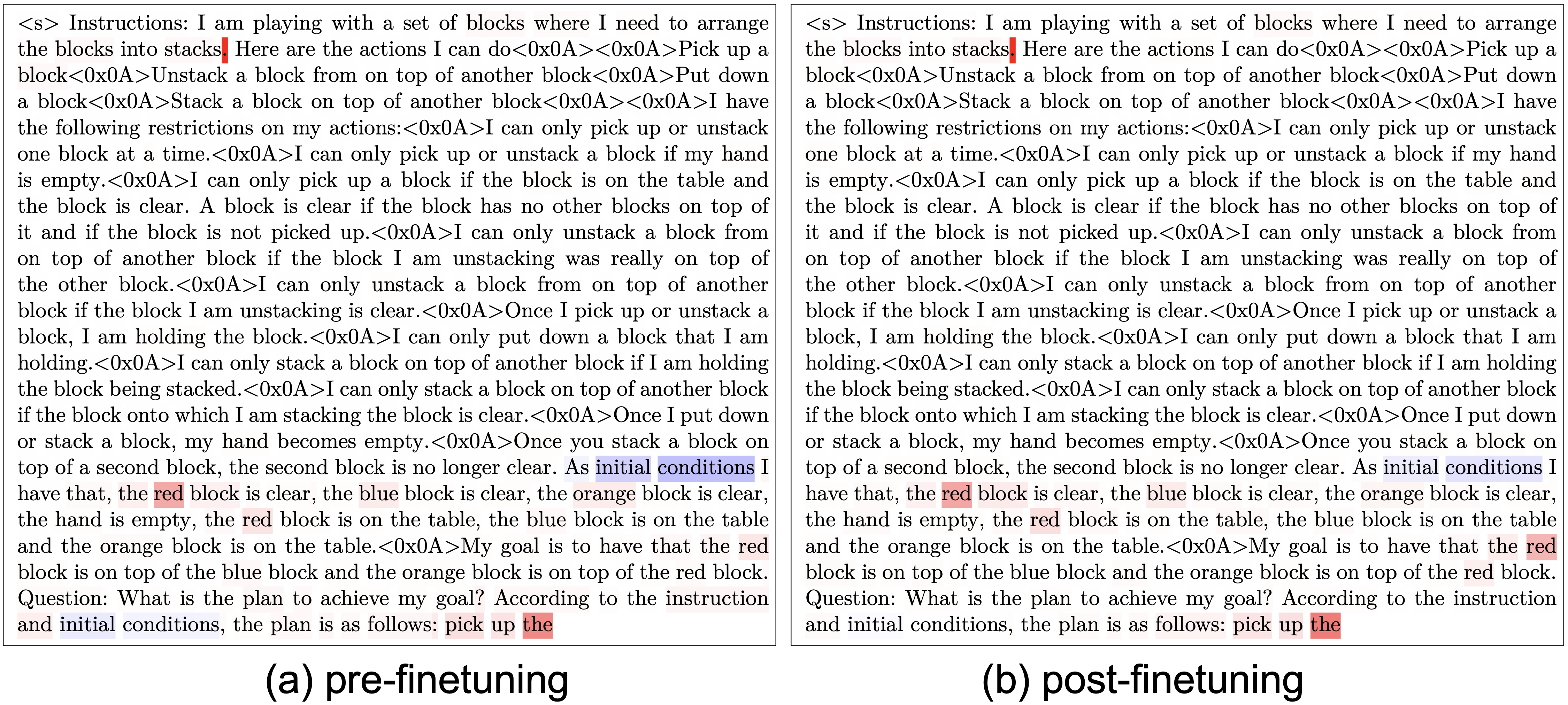}
  \caption{Attention visualization. The correct next token in the example is ``red". The normalized attention on token `red' for the pre-finetuning model is sum$\left['red'\right]=0.55$ and for the post-finetuning model is sum$\left['red'\right]=0.85$.}
  \label{fig:attention_visualization_bw}
\end{figure}

Beyond using task solved rates as a primary metric for evaluating model performance, we also analyze the attention mechanisms of transformers, offering a visual comparison between pre-finetuning and post-finetuning models. This analysis utilizes the Layer-wise Relevance Propagation (LRP,~\cite{achtibat2024attnlrp}) method, which integrates attention layers to provide a more detailed interpretation of the performance differences between models. An example of this visualization is shown in Figure \ref{fig:attention_visualization_bw}. It is evident that after fine-tuning, LLMs focus more on the correct tokens, thereby increasing the likelihood of making correct decisions from the very first step.

We then investigate how the model's attention is affected by fine-tuning on datasets selected through different methods, namely Random, CMDS-$l$, and CMDS-$g$. We randomly selected 10 examples from the hold-out test tasks, with the results summarized in Table \ref{table:attention_next_word}. Among these examples, CMDS-$g$ enabled the model to achieve the highest attention on the correct next token in 8 out of 10 test cases, while CMDS-$l$ led in 2 out of 10 cases. These outcomes are consistent with the model's solved rate performance reported in previous sections. This attention visualization serves as additional evidence supporting the effectiveness of the fine-tuning process.
\begin{table}[th]
  \centering
  \begin{tabular}{ccccc}
    \toprule
    & & \multicolumn{3}{c}{\textbf{Post-Finetuning}} \\
    \cmidrule(lr){3-5}
    \textbf{Correct Next Token} & \textbf{Pre-Finetuning} & \textbf{Random} & \textbf{CMDS-$l$} & \textbf{CMDS-$g$} \\
    \midrule
    sum$\left['red'\right]$    & 0.554 & 0.845 & 0.873 & \textbf{0.928} \\
    sum$\left['blue'\right]$   & 0.880 & 0.937 & \textbf{0.976} & 0.967 \\
    sum$\left['red'\right]$    & 0.602 & 0.881 & 0.936 & \textbf{1.02} \\
    sum$\left['blue'\right]$   & 0.936 & 0.944 & 0.953 & \textbf{0.998} \\
    sum$\left['red'\right]$    & 0.596 & 0.944 & 0.987 & \textbf{1.053} \\
    sum$\left['yellow'\right]$   & 0.910 & 0.954 & 0.957 & \textbf{0.980} \\
    sum$\left['yellow'\right]$   & 0.234 & 0.786 & 0.862 & \textbf{0.872}\\
    sum$\left['orange'\right]$   & 0.317 & 1.117 & \textbf{1.166} & 1.128 \\
    sum$\left['orange'\right]$   & 0.319 & 1.180 & 1.172 & \textbf{1.187} \\
    sum$\left['blue'\right]$     & 0.217 & 0.913 & 0.945 & \textbf{0.966} \\
    \bottomrule
  \end{tabular}
  \caption{Attention on the correct next word for each sample index before and after fine-tuning. Base model is Llama-2-7b in this experiment.}
  \label{table:attention_next_word}
\end{table}

\clearpage
\section{Domain Details and Prompts} \label{app:domain_details}
Our implementations are based on the codebase from ~\cite{valmeekam2024planning}. In this section, we provide the domain properties, instructions, actions, predicates, and prompts used in our experiments.

\subsection{Blocksworld Domain}

\subsubsection{Domain Properties}
The Blocksworld domain focuses on stacking blocks on a table. One hand is available to move blocks, and only one block may be moved by the hand at a time. Blocks cannot be moved if there are blocks on top of them, and blocks cannot be stacked on a block that already has another block on top of it. The goals specify the order in which blocks within a stack should be stacked but may include multiple stacks or request that blocks be left on the table. We adopt the 4-operator version of the classic Blocksworld.

Below is the domain instruction included in the prompts to the LLMs. For one-shot prompting, the prompt contains three components: the domain instruction, an example task and its solution, the query task. For zero-shot prompting, the prompt contains two components: the domain instruction and the query task. 

\begin{tcolorbox}[colback=gray!10, colframe=black, title=Domain Instruction]
I am playing with a set of blocks where I need to arrange the blocks into stacks. Here are the actions I can do:\\

Pick up a block\\
Unstack a block from on top of another block\\
Put down a block\\
Stack a block on top of another block\\

I have the following restrictions on my actions:\\
I can only pick up or unstack one block at a time.\\
I can only pick up or unstack a block if my hand is empty.\\
I can only pick up a block if the block is on the table and the block is clear.\\
A block is clear if the block has no other blocks on top of it and if the block is not picked up.\\
I can only unstack a block from on top of another block if the block I am unstacking was really on top of the other block.\\
I can only unstack a block from on top of another block if the block I am unstacking is clear.\\
Once I pick up or unstack a block, I am holding the block.\\
I can only put down a block that I am holding.\\
I can only stack a block on top of another block if I am holding the block being stacked.\\
I can only stack a block on top of another block if the block onto which I am stacking the block is clear.\\
Once I put down or stack a block, my hand becomes empty.\\
Once you stack a block on top of a second block, the second block is no longer clear.
\end{tcolorbox}

\begin{tcolorbox}[colback=gray!10, colframe=black, title=Actions]
\begin{enumerate}
    \item \textbf{pick-up}: pick up the $\{\}$.
    \item \textbf{put-down}: put down the $\{\}$.
    \item \textbf{stack}: stack the $\{\}$ on top of the $\{\}$.
    \item \textbf{unstack}: unstack the $\{\}$ from on top of the $\{\}$.
\end{enumerate}
\end{tcolorbox}

\begin{tcolorbox}[colback=gray!10, colframe=black, title=Predicates]
\begin{enumerate}
    \item \textbf{ontable}: the $\{\}$ is on the table.
    \item \textbf{clear}: the $\{\}$ is clear.
    \item \textbf{handempty}: the hand is empty.
    \item \textbf{holding}: the hand is currently holding $\{\}$.
    \item \textbf{on}: the $\{\}$ is on top of the $\{\}$.
\end{enumerate}
\end{tcolorbox}

\begin{tcolorbox}[colback=gray!10, colframe=black, title=An Example Task and Its Solution in the Blocksworld Domain]
\texttt{[STATEMENT]}\\
As initial conditions I have that, the blue block is clear, the hand is empty, the blue block is on top of the yellow block, the orange block is on top of the red block, the yellow block is on top of the orange block and the red block is on the table. \\
My goal is to have that the red block is on top of the blue block, the blue block is on top of the orange block and the yellow block is on top of the red block.\\

My plan is as follows:\\
\texttt{[PLAN]}

unstack the blue block from on top of the yellow block\\
put down the blue block\\
unstack the yellow block from on top of the orange block\\
put down the yellow block\\
unstack the orange block from on top of the red block\\
put down the orange block\\
pick up the blue block\\
stack the blue block on top of the orange block\\
pick up the red block\\
stack the red block on top of the blue block\\
pick up the yellow block\\
stack the yellow block on top of the red block\\
\texttt{[PLAN END]}
\end{tcolorbox}

\begin{tcolorbox}[colback=gray!10, colframe=black, title=An Example Query in the Blocksworld Domain]
\texttt{[STATEMENT]}\\
As initial conditions I have that, the yellow block is clear, the hand is empty, the red block is on top of the blue block, the blue block is on top of the orange block, the yellow block is on top of the red block and the orange block is on the table.\\
My goal is to have that the blue block is on top of the orange block, the orange block is on top of the red block and the yellow block is on top of the blue block.\\

My plan is as follows:\\
\texttt{[PLAN END]}
\end{tcolorbox}

\subsubsection{Graph Encoding}
After representing the planning tasks using their graph structures, we proceed to encode these graphs into vectors. Below, we describe the graph encoding function employed for the Blocksworld domain in this paper. It's important to note that there are various methods to perform graph encoding, and our approach is just one of many possibilities. Drawing inspiration from \cite{rivlin2020generalized, silver2021planning}, we designed our own graph encoding function.

\begin{figure}[ht]
  \centering
  \includegraphics[width=1.0\linewidth]{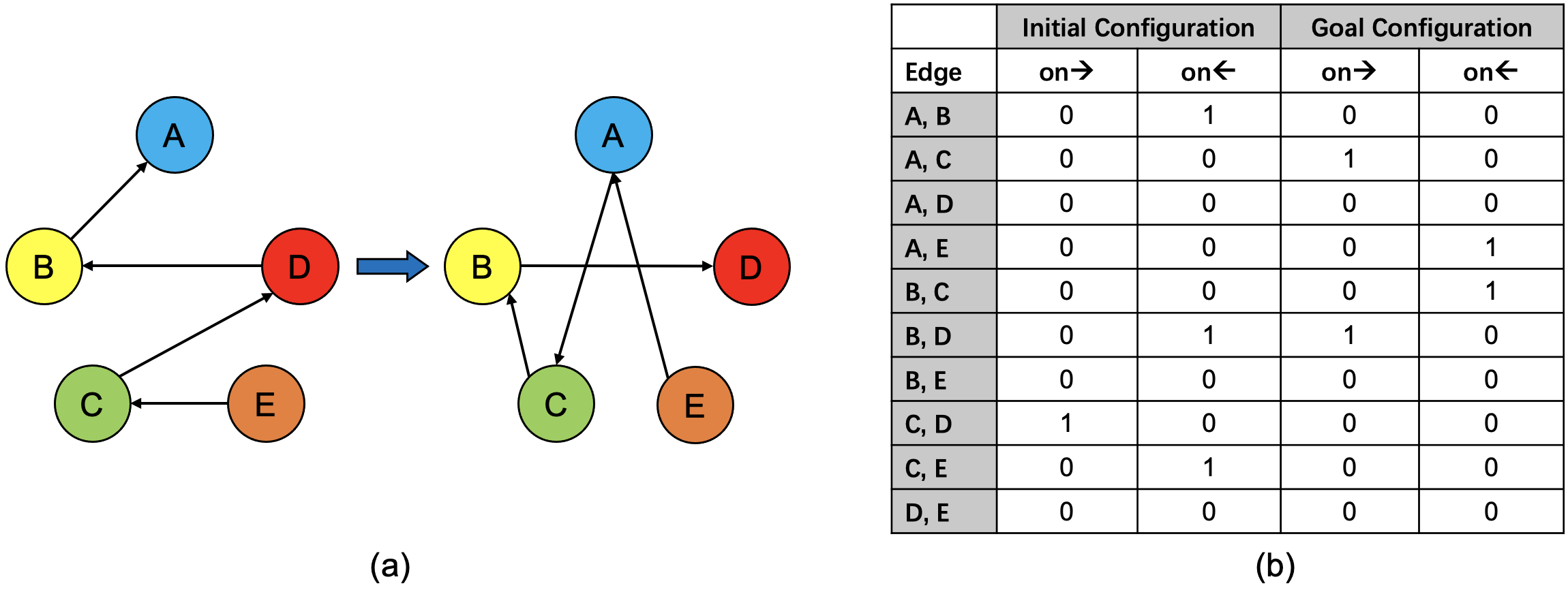}
  \caption{Encoding graphs into vectors. (a) An example task in Blocksworld. (b) Vector encoding of the example task.}
  \label{fig:graph_embedding_bw}
\end{figure}

As illustrated in Figure \ref{fig:graph_embedding_bw}, our encoding converts the graph representation into a 10-dimensional vector, where a value of 1 represents a directed edge between two nodes and a value of 0 indicates no edge between the nodes. In this manner, both the initial configuration and the goal configuration are encoded as 10-dimensional vectors. To encapsulate the complete task information from both the initial and goal configurations, we concatenate these two vectors, resulting in the full vector representation of the planning task. This process converts a planning task from a natural language description to a graph representation, and finally, to a vector encoding.

Figure \ref{fig:graph_embedding_bw} provides the encoding method for 5-block tasks. For Blocksworld tasks with fewer blocks, we pad the non-existing nodes and edges with 0 and then add 1 to all elements in the vector. Thus, non-existing nodes and edges will have a value of 0, nodes that exist but have no edge between them will have a value of 1, and nodes with an existing edge will have a value of 2.

We have omitted the specific graph encoding method for the Logistics domain here, as it functions similarly to the approach shown in Figure \ref{fig:graph_embedding_bw}. Readers are encouraged to design their own encoding functions tailored to their specific needs.

\subsubsection{Dataset Generation}
In this section, we outline the process used to generate the dataset of planning tasks, as well as the procedures for splitting and preparing the training and testing data.

For the motivation example, which requires a large number of instances, we employed a randomization method with uniqueness checking to generate 50,000 distinct instances. For other experiments, we employ randomization to generate 128, 5000, 5000, and 5000 instances for 3-block, 4-block, 5-block, and 6-block tasks, respectively. The initial and goal configurations are randomly generated, and we ensure that each task in a dataset is distinct. For research question 1, we randomly select 100, 500, 500, and 500 tasks from each group as the test data for each task setting. Note that the test data are separated from the training data. For test data in research question 2, we uniformly sample from the four types of task settings to collectively create a testing dataset with 1000 instances, as detailed performance on different types of tasks was unnecessary for this analysis. For train data in research question 2, we sample from the remaining data to create the fine-tuning datasets, either according to CMDS algorithms or Random.

\subsection{Logistics Domain}
In the Logistics domain, the goal is to transport the packages to designated locations with trucks and airplanes, where trucks can only move between locations within the same city and airplanes can only fly between cities. Locations are grouped by cities. Note that there are no restrictions on the positions of the trucks and airplanes in the goal configuration. Similarly, we provide an example task and its solution from the Logistics domain below. 

\begin{tcolorbox}[colback=gray!10, colframe=black, title=Domain Instruction]
I have to plan logistics to transport packages within cities via trucks and between cities via airplanes. Locations within a city are directly connected (trucks can move between any two such locations), and so are the cities. In each city, there is exactly one truck and each city has one location that serves as an airport.\\

Here are the actions that can be performed:\\
Load a package into a truck. \\
Load a package into an airplane.\\
Unload a package from a truck. \\
Unload a package from an airplane.\\ 
Drive a truck from one location to another location. \\
Fly an airplane from one city to another city.\\

The following are the restrictions on the actions:\\
A package can be loaded into a truck only if the package and the truck are in the same location.\\
Once a package is loaded into a truck, the package is not at the location and is in the truck.   \\
A package can be loaded into an airplane only if the package and the airplane are in the same location.\\
Once a package is loaded into an airplane, the package is not at the location and is in the airplane.\\
A package can be unloaded from a truck only if the package is in the truck.\\
Once a package is unloaded from a truck, the package is not in the truck and is at the location of the truck.\\
A package can be unloaded from an airplane only if the package in the airplane.\\
Once a package is unloaded from an airplane, the package is not in the airplane and is at the location of the airplane.   \\
A truck can be driven from one location to another if the truck is at the from-location and both from-location and to-location are locations in the same city.\\
Once a truck is driven from one location to another, it is not at the from-location and is at the to-location.\\
An airplane can be flown from one city to another if the from-location and the to-location are airports and the airplane is at the from-location.\\
Once an airplane is flown from one city to another the airplane is not at the from-location and is at the to-location.
\end{tcolorbox}

\begin{tcolorbox}[colback=gray!10, colframe=black, title=Actions]
\begin{enumerate}
\item \textbf{load-truck}: load $\{\}$ into $\{\}$ at $\{\}$.
\item \textbf{load-airplane}: load $\{\}$ into $\{\}$ at $\{\}$.
\item \textbf{unload-truck}: unload $\{\}$ from $\{\}$ at $\{\}$.
\item \textbf{unload-airplane}: unload $\{\}$ from $\{\}$ at $\{\}$.
\item \textbf{drive-truck}: drive $\{\}$ from $\{\}$ to $\{\}$ in $\{\}$.
\item \textbf{fly-airplane}: fly $\{\}$ from $\{\}$ to $\{\}$.
\item \textbf{drive-truck }: unstack the $\{\}$ from on top of the $\{\}$.
\end{enumerate}
\end{tcolorbox}

\begin{tcolorbox}[colback=gray!10, colframe=black, title=Predicates]
\begin{enumerate}
\item \textbf{airport}: $\{\}$ is an airport.
\item \textbf{at}: $\{\}$ is at $\{\}$.
\item \textbf{in}: $\{\}$ is in $\{\}$.
\item \textbf{in-city}: $\{\}$ is in the city $\{\}$.
\end{enumerate}
\end{tcolorbox}

\begin{tcolorbox}[colback=gray!10, colframe=black, title=An Example Task and Its Solution in the Logistics Domain]
\texttt{[STATEMENT]}\\
As initial conditions I have that, location\_0\_0 is an airport, location\_1\_0 is an airport, airplane\_0 is at location\_0\_0, package\_0 is at location\_1\_1, truck\_0 is at location\_0\_1, truck\_1 is at location\_1\_0, location\_0\_0 is in the city city\_0, location\_0\_1 is in the city city\_0, location\_1\_0 is in the city city\_1 and location\_1\_1 is in the city city\_1.\\

My goal is to have that package\_0 is at location\_0\_0.\\

My plan is as follows:\\
\texttt{[PLAN]}\\
drive truck\_1 from location\_1\_0 to location\_1\_1 in city\_1\\
load package\_0 into truck\_1 at location\_1\_1\\
drive truck\_1 from location\_1\_1 to location\_1\_0 in city\_1\\
unload package\_0 from truck\_1 at location\_1\_0\\
fly airplane\_0 from location\_0\_0 to location\_1\_0\\
load package\_0 into airplane\_0 at location\_1\_0\\
fly airplane\_0 from location\_1\_0 to location\_0\_0\\
unload package\_0 from airplane\_0 at location\_0\_0\\
\texttt{[PLAN END]}
\end{tcolorbox}

\begin{tcolorbox}[colback=gray!10, colframe=black, title=An Example Query in the Logistics Domain]
\texttt{[STATEMENT]} \\
As initial conditions I have that, location\_0\_0 is an airport, location\_1\_0 is an airport, airplane\_0 is at location\_1\_0, package\_0 is at location\_1\_1, truck\_0 is at location\_0\_1, truck\_1 is at location\_1\_1, location\_0\_0 is in the city city\_0, location\_0\_1 is in the city city\_0, location\_1\_0 is in the city city\_1 and location\_1\_1 is in the city city\_1.\\

My goal is to have that package\_0 is at location\_0\_0.\\

My plan is as follows:\\
\texttt{[PLAN]}
\end{tcolorbox}

\subsection{Dataset Generation}
In Logistics, four variables—number of cities, locations within a city, airplanes, and packages—affect task complexity. However, determining which of these variables has the greatest impact on task complexity and difficulty is challenging. To address this, we generated a dataset of six thousand instances with varying values for these variables. The ranges for the variables are as follows: number of cities: [2], number of locations: [2, 3], number of airplanes: [1, 2], and number of packages: [1, 2]. From this dataset, we randomly sampled three hundred instances as the test data, while the remaining instances were used as the training dataset. The test data were kept separate from the training data to ensure a clear evaluation.

\end{document}